\documentclass[sigconf]{acmart}
\usepackage[ruled,vlined]{algorithm2e}
\usepackage{multirow}
\usepackage{booktabs}
\usepackage{xspace}
\usepackage{mathtools}
\usepackage{enumitem}
\usepackage{hyperref}
\usepackage{amsthm}

\usepackage{threeparttable}
\usepackage{amsmath}
\usepackage{bbding}
\usepackage{pifont}
\usepackage{ulem}
\usepackage{threeparttable} 
\definecolor{lightblue}{RGB}{68,114,196}  

\AtBeginDocument{%
  \providecommand\BibTeX{{%
    \normalfont B\kern-0.5em{\scshape i\kern-0.25em b}\kern-0.8em\TeX}}}

\setcopyright{acmcopyright}
\copyrightyear{2026}
\acmYear{2026}
\acmDOI{XXXXXXX.XXXXXXX}

\acmPrice{15.00}
\acmISBN{978-1-4503-XXXX-X/18/06}

\setcopyright{none}
\renewcommand\footnotetextcopyrightpermission[1]{}
\begin{document}

\title{Effective and Efficient Cross-City Traffic Knowledge Transfer: A Privacy-Preserving Perspective}

\author{Zhihao Zeng}
\affiliation{%
  \institution{Zhejiang University}
}
\email{zengzhihao@zju.edu.cn}

\author{Ziquan Fang}
\affiliation{%
  \institution{Zhejiang University}
}
\email{zqfang@zju.edu.cn}

\author{Yuting Huang}
\affiliation{%
  \institution{Zhejiang University}
}
\email{huangyuting@zju.edu.cn}

\author{Lu Chen}
\affiliation{%
  \institution{Zhejiang University}
}
\email{luchen@zju.edu.cn}

\author{Yunjun Gao}
\affiliation{%
  \institution{Zhejiang University}
}
\email{gaoyj@zju.edu.cn}

\begin{abstract}

Traffic prediction (TP) is a core task in urban computing, aiming to forecast future traffic conditions from historical observations. To overcome the scarcity of traffic data in emerging cities, recent studies have explored Federated Traffic Knowledge Transfer (FTT), which leverages data-rich source cities to assist data-scarce target cities without raw data sharing. However, existing FTT approaches are limited by three unresolved challenges: (i) potential \textit{privacy leakage} since gradients or parameters generated during federated computing can still be inverted, (ii) severe \textit{cross-city distribution discrepancies} that reduce transfer effectiveness, and (iii) \textit{low data quality} caused by missing or unreliable sensor readings. To address these challenges, we propose \textbf{FedTT}, a novel federated framework for cross-city traffic knowledge transfer with privacy-preserving. FedTT introduces three innovations: (i) a lightweight \textbf{Traffic Secret Aggregation (TSA)} protocol that achieves secure knowledge aggregation without sacrificing efficiency or accuracy; (ii) a \textbf{Traffic Domain Adapter (TDA)} that explicitly aligns heterogeneous source–target distributions for more effective transfer, and (iii) a \textbf{Traffic View Imputation (TVI)} method that leverages spatio-temporal dependencies to complete missing traffic data robustly. Extensive experiments on four real-world datasets show that FedTT achieves significant improvements over 18 state-of-the-art baselines, consistently reducing prediction error while maintaining strong privacy protection.

\end{abstract}




\maketitle

\section{Introduction}
\label{sec:intro}

The substantial rise in population density and vehicle volume has presented significant challenges to urban transportation. As a pivotal solution, \textbf{Traffic Prediction} (TP)~\cite{YangCHWXZT24, jyq24, zys23}, leverage widespread sensors in the road network to forecast traffic conditions based on historical traffic data (e.g. traffic flow, speed, and occupancy), which not only facilitates the effective allocation of public transportation resources~\cite{czm21} but also contributes to alleviating traffic congestion~\cite{YuanCZZYS22}, addressing key issues in urban transportation.

In the spatio-temporal databases, numerous TP methods have been proposed~\cite{zys23,jjh23,jjw23}, which typically rely on a large number of labeled data to train high-performing traffic models. However, urban traffic data is often insufficient or unavailable~\cite{LinXLZ18, WangGMLY19, WangMLC22}, particularly in emerging cities, such as developing regions in Midwestern United States~\cite{us}, where sensors are newly deployed or data collection is still in its early stages. In such cases, training traffic models becomes particularly challenging and prone to overfitting, limiting the accuracy of TP tasks~\cite{JinCY23, MoG23}. To address this challenge, \textbf{Transfer Learning} (TL)~\cite{st18, smd19} has been widely adopted to enhance the traffic modeling capabilities of the data-scarce target city by transferring traffic knowledge from data-rich source cities.


\begin{figure}[t]
    \centering
    \includegraphics[width=0.47\textwidth]{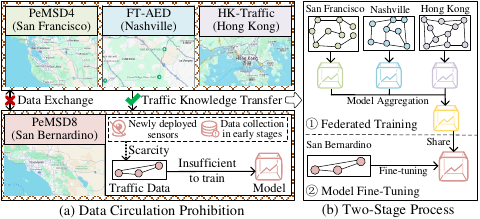}
    \vspace{-4mm}
    \caption{Privacy-Preserving Traffic Knowledge Transfer}
    \label{fig:intro1}
    \vspace{-4mm}
\end{figure}

\begin{figure*}[t]
    \centering
    \vspace{-4mm}
    \includegraphics[width=\textwidth]{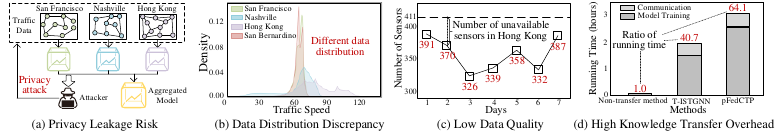}
    \vspace{-6mm}
    \caption{Four Unresolved Challenges of Existing Federated Traffic Knowledge Transfer (FTT) Methods}
    \label{fig:intro2}
    \vspace{-4mm}
\end{figure*}
To enable data-scarce target cities to acquire diverse traffic knowledge and improve the robustness of trained models, existing TL-based TP methods~\cite{LiuZY23,0005G0YFW22, TangQCLWM22} tend to perform joint traffic knowledge transfer from multiple source cities. However, these approaches typically rely on centralized frameworks, which involve sharing and exchanging traffic data among cities without considering data privacy. These methods assume that different cities can directly share their original traffic data, which is overly idealistic in reality. This is because, the direct sharing of city traffic data risks privacy leakage~\cite{LiuZZY20, czm21, YangCHWXZT24} due to the presence of personal and sensitive information, such as license plate numbers and travel trajectories. Moreover, with growing privacy concerns, many laws and regulations, such as GDPR~\cite{gdpr} and CCPA~\cite{ccpa}, mandate data collectors to minimize non-essential data transmission and avoid centralized storage of private information. Therefore, maintaining the decentralization of traffic data in TP is critical to safeguarding privacy. As shown in Fig.~\ref{fig:intro1}(a), PeMSD4~\cite{pems}, FT-AED~\cite{ftaed}, HK-Traffic~\cite{hk}, and PeMSD8~\cite{pems} are four commonly used traffic datasets, which correspond to San Francisco, Nashville, Hong Kong, and San Bernardino, respectively. Among these, San Francisco, Nashville, and Hong Kong represent the source cities, while San Bernardino serves as the target city. Due to legal restrictions\textcolor{blue}{(e.g., GDPR~\cite{gdpr} and CCPA~\cite{ccpa})}, traffic data cannot be shared or exchanged among cities, meaning each city can only access its local data. Under these constraints, transferring traffic knowledge from these three source cities to the target city without sharing and exchanging raw traffic data becomes challenging and critical.

\textbf{Federated Learning} (FL)~\cite{yxw24, byl24, YangCHWXZT24}, a privacy-preserving distributed learning paradigm, has been widely used in numerous applications to address privacy concerns such as urban computing~\cite{ysw22} and transportation management~\cite{YangCHWXZT24}. For instance, JD Company (one of the largest e-commerce companies in China) develops the Fedlearn platform to help protect data privacy for TP applications~\cite{bg20}. To protect data privacy, FL uses multiple clients to jointly train models by sharing only intermediate results such as training gradients or model parameters without local/raw data exchange. 
Inspired by the success of federated learning, recent studies, including T-ISTGNN~\cite{QiWBLYA23} and pFedCTP~\cite{Zhang00XLC24}, have explored utilizing the FL framework to transfer traffic knowledge while preserving data privacy, a new problem referred to as \textbf{Federated Traffic Knowledge Transfer} (FTT). As illustrated in Fig.~\ref{fig:intro1}(b), these methods typically follow a two-stage process. \underline{In the first stage}, the three source cities (i.e., San Francisco, Nashville, and Hong Kong) use their local traffic data to train individual local models. Subsequently, the cities upload training gradients or model parameters to a central server, which aggregates them to produce a global traffic model. The server then broadcasts the global model back to the source cities for local model updates, iterating this process until the global model converges. \underline{In the second stage}, the converged global model is shared with the target city (i.e., San Bernardino) to initialize its traffic model, which is further refined through training and fine-tuning using the target city's local traffic data. While this two-stage framework has become the mainstream approach in FTT, it faces four unresolved challenges—privacy, effectiveness, robustness, and efficiency—that hinder its application in real-world traffic knowledge transfer scenarios, as illustrated in Fig.~\ref{fig:intro2}.

\textbf{\textit{Challenge 1: How to effectively protect data privacy during federated traffic knowledge transfer?}} Although existing methods utilize federated learning to avoid direct data exchange, there is still a potential risk of data privacy leakage. This arises because these methods require the uploading of training gradients or model parameters for aggregation during federated training, which may allow attackers to infer raw data by inference attacks~\cite{kg24,ybw24,llz24}, as depicted in Fig.~\ref{fig:intro2}(a). To mitigate this risk, a straightforward approach is to apply privacy-preserving techniques such as Homomorphic Encryption (HE)~\cite{rlr78} and Differential Privacy (DP) for secure aggregation on the uploaded data during federated training. However, HE could introduce significant additional computation and communication overheads, which diminishes training efficiency, while DP lowers data utility and thus decreases model accuracy, as proved by previous studies~\cite{phw24,ott23,wd24}. Therefore, ensuring the protection of traffic data privacy without compromising training efficiency and model accuracy remains a significant challenge.

\textbf{\textit{Challenge 2: How to mitigate the impact of cross-city traffic data distribution discrepancies on federated traffic knowledge transfer?}} None of the previous federated traffic knowledge transfer (FTT) studies have considered the discrepancies in traffic data distribution between source and target cities, which decreases the effectiveness of traffic knowledge transfer~\cite{LiuZY23, 0005G0YFW22, TangQCLWM22}. Specifically, the traffic domain varies significantly across cities, with distinct distributions of traffic flow, speed, and occupancy data. As shown in Fig.~\ref{fig:intro2}(b), we illustrate the frequency density distribution of traffic speed data for San Francisco, Nashville, Hong Kong, and San Bernardino. As observed, the data distributions of San Francisco and San Bernardino are very similar, suggesting that their traffic domains are closely related. In contrast, the data distributions of Nashville and San Bernardino differ significantly, indicating that their traffic domains are quite distinct. Consequently, traffic knowledge transfer from San Francisco to San Bernardino results in smaller prediction errors and is more effective than transfer from Nashville to San Bernardino.
Overall, how to address traffic domain discrepancies between the source and target cities to improve the performance of FTT is the second challenge that must be addressed.

\textbf{\textit{Challenge 3: How to overcome low traffic data quality to improve federated traffic knowledge transfer?}} Existing federated traffic knowledge transfer methods assume that traffic data quality is consistently good and reliable, failing to account for cases of missing traffic data, which is actually common in real-life traffic scenarios. As shown in Fig.~\ref{fig:intro2}(c), we illustrate the number of available sensors over a week in HK-Traffic, which has 411 sensors in total. Due to sensor failures or updates~\cite{hty24, hlq21}, the number of available sensors may fluctuate over time, disrupting the model training process. In this context, existing methods directly default to zero values for missing traffic data, which degrades model accuracy in traffic knowledge transfer. 
Therefore, enhancing the traffic data quality to improve the robustness of FTT is another challenge.

\textbf{\textit{Challenge 4: How to reduce computation and communication overheads during federated traffic knowledge transfer?}}
Existing federated traffic knowledge transfer approaches are inefficient due to several factors. First, their two-stage knowledge transfer process operates serially, which is inefficient. Second, to achieve good prediction accuracy, existing traffic prediction models are designed with high complexity and a large number of parameters~\cite{zys23,jjh23,jjw23}, leading to significant computation and communication overheads, further limiting the efficiency of traffic knowledge transfer. As illustrated in Fig.~\ref{fig:intro2}(d), we report the running time (in hours) of the SOTA FTT methods (i.e., T-ISTGNN and pFedCTP) and a non-transfer method (i.e., Gated Recurrent Unit (GRU)~\cite{Chung14} model is trained directly on the target city's data without traffic knowledge transfer). Compared to the non-transfer method, the existing FTT methods exhibit considerably higher computation and communication overheads, with running times exceeding 40$\times$ that of the non-transfer method. As a result, how to reduce computation and communication overheads in traffic knowledge transfer to improve the efficiency of FTT is the fourth unaddressed challenge.

To address the above four challenges, we propose FedTT, an effective, efficient, and privacy-aware \underline{Fed}erated \underline{T}raffic knowledge \underline{T}ransfer framework. Unlike existing federated traffic knowledge transfer methods that rely on federated training and model fine-tuning, FedTT focuses on transforming the traffic data from the source cities' traffic domain to the target city's traffic domain and training the target city's traffic model on the transformed data. To address \textbf{\textit{Challenge 1}}, FedTT introduces the Traffic Secret Transmission (TST) method to aggregate the transformed data from source cities. It also designs a lightweight traffic secret aggregation approach to protect traffic data privacy during the data aggregation process. To overcome \textbf{\textit{Challenge 2}}, FedTT develops the Traffic Domain Adapter (TDA), which uniformly transforms the data from the traffic domain of source cities to that of the target city through traffic domain transformation, alignment, and classification operations. 
To deal with \textbf{\textit{Challenge 3}}, FedTT introduces the Traffic View Imputation (TVI) method to complete and predict missing traffic data by capturing the spatio-temporal dependencies of traffic views. To tackle \textbf{\textit{Challenge 4}}, FedTT introduces Federated Parallel Training (FPT), which enables the simultaneous training of different modules. Specifically, it employs split learning and parallel optimization to decompose the training process and reduce data transmission. In summary, this paper makes the following key contributions.

\vspace{-1mm}
\begin{itemize}[leftmargin=*]
\item{}
We propose FedTT, an effective, efficient, and privacy-aware cross-city traffic knowledge transfer framework designed to overcome the limitations of privacy leakage risk, cross-city data discrepancy, low data quality, and inefficient knowledge transfer.

\item{}
To enhance traffic data quality, we define traffic view and design a Traffic View Imputation (TVI) module that captures the spatio-temporal dependencies through spatial view extension and temporal view enhancement, enabling robust traffic knowledge transfer even under low traffic data quality (Section 4.1).

\item{}
To mitigate the impact of traffic data distribution discrepancies on federated traffic knowledge transfer, we employ a Traffic Domain Adapter (TDA) module to uniformly transform the traffic data from source cities' traffic domains to that of the target city, enabling effective traffic knowledge transfer (Section 4.2).

\item{}
To effectively protect traffic data privacy, we propose a Traffic Secret Transmission (TST) module to securely aggregate the transformed data by a lightweight secret aggregation method without compromising training efficiency or model accuracy, enabling privacy-preserving knowledge transfer (Section 4.3).

\item{}
To improve training efficiency, we introduce a Federated Parallel Training (FPT) module to reduce communication overhead and improve training parallelism through split learning and parallel optimization, enabling the simultaneous training of modules and efficient traffic knowledge transfer (Section 4.4).

\item{}
Extensive experiments conducted on 4 real-world datasets demonstrate that FedTT achieves state-of-the-art performance in 3 traffic prediction tasks, with a reduction in MAE ranging from \textbf{5.43\% to 22.78\%}, compared to 14 baseline methods (Section 5).





\end{itemize}

\vspace{-2mm}
\section{Related Work}
\vspace{-1mm}
\label{sec:related}

\subsection{Traffic Prediction}
Traffic prediction plays a critical role in the development of smart cities and has garnered significant attention in the spatio-temporal data mining community. Currently, deep learning techniques~\cite{rf58} are widely employed in traffic prediction tasks. Convolutional models, such as Convolutional Neural Networks (CNN)~\cite{LeCun98} and Graph Convolutional Networks (GCN)~\cite{Kipf17}, are used to capture spatial correlations in traffic time-series data. Meanwhile, sequential models including Gated Recurrent Units (GRU)~\cite{Chung14} and Long Short-Term Memory (LSTM)~\cite{Gers00}, are employed to extract temporal dependencies from the data. Several advanced models have achieved state-of-the-art performance. For instance, ST-SSL~\cite{jjh23} improves traffic pattern representation to account for spatial and temporal heterogeneity through a self-supervised learning framework. DyHSL~\cite{zys23} leverages hypergraph structure information to model the dynamics of a traffic network, updating the representation of each node by aggregating messages from associated hyperedges. Additionally, PDFormer~\cite{jjw23} introduces a spatial self-attention module to capture dynamic spatial dependencies and a flow-delay-aware feature transformation module to model the time delays in spatial information propagation. Since this paper is not intended to propose another more complex prediction model, a detailed analysis of existing traffic prediction models can be found in surveys~\cite{gyj24,sk24}. However, these models are centralized and rely on traffic data uploads from sensors to a central server, which poses a risk of data leakage.


To address data privacy concerns, several traffic prediction studies~\cite{YuanCZZYS22, LaiTWH23, XiaJC23, LiL24a, YangCHWXZT24, LiuSLWG24} in federated environments have been proposed. Specifically, FedGRU~\cite{LiuZZY20} pioneers the integration of GRU into FL for TP tasks, employing federated averaging to aggregate models and a joint announcement protocol to enhance model scalability. Subsequently, CNFGNN~\cite{czm21} separates the modeling of temporal dynamics on the device from spatial dynamics on the server, using alternating optimizations to reduce communication costs and facilitate computation on edge devices.
Moreover, FedGTP~\cite{YangCHWXZT24} promotes the adaptive exploitation of inter-client spatial dependencies to enhance prediction performance while ensuring data privacy. However, urban traffic data is often insufficient or unavailable, particularly in emerging cities. Training traffic models in these data-scarce cities is prone to overfitting, which undermines model performance and affects the accuracy of TP tasks.
\textbf{\textit{In contrast, we aim to propose a federated traffic prediction framework that efficiently transfers traffic knowledge from data-rich cities to data-scarce cities, enhancing TP capabilities for the latter.}}

\subsection{Traffic Knowledge Transfer}
Transfer learning can enhance the traffic model capabilities of data-scarce target cities by transferring traffic knowledge from data-rich source cities in traffic prediction tasks. Existing studies can be broadly categorized into three types: Single-Source Traffic Knowledge Transfer (STT), Multi-Source Traffic Knowledge Transfer (MTT), and Federated Traffic Knowledge Transfer (FTT), in chronological order from earliest to most recent.

First, STT~\cite{Jin0022, FangWPCG22, WangGMLY19, LinXLZ18, YaoXLWZ23, HuangSZZY23, LiTM23, ChenGZLS22, WangMLC22} studies focus on transferring traffic knowledge from a single source city to a target city. Specifically, TransGTR~\cite{JinCY23} jointly learns transferable structure generators and forecasting models across cities to enhance prediction performance in data-scarce target cities. Next, CityTrans~\cite{OuyangYZZWH24} leverages adaptive spatio-temporal knowledge and domain-invariant features for accurate traffic prediction in data-scarce cities. Additionally, MGAT~\cite{MoG23} uses a meta-learning algorithm to extract multi-granular regional features from each source city to improve the effectiveness of traffic knowledge transfer. However, the performance of these STT methods can be significantly compromised when there are substantial differences in traffic data distribution between the source and target cities.

Second, MTT~\cite{YaoLWTL19, LiuGZZCZZSY21, Zhang00XLC24, LZ022} studies the joint transfer of traffic knowledge from multiple source cities to a target city, enabling the target city to acquire diverse traffic knowledge and enhancing the robustness of the trained traffic models. Specifically, TPB~\cite{LiuZY23} uses a traffic patch encoder to create a traffic pattern bank, which data-scarce cities query to establish relationships, aggregate meta-knowledge, and construct adjacency matrices for future traffic prediction. Next, ST-GFSL~\cite{0005G0YFW22} transfers knowledge through parameter matching to retrieve similar spatio-temporal features and defines graph reconstruction loss to guide structure-aware learning. Additionally, DastNet~\cite{TangQCLWM22} employs graph representation learning and domain adaptation techniques to create domain-invariant embeddings for traffic data. However, these methods rely on centralized frameworks, which involves sharing and exchanging traffic data across cities without considering traffic data privacy.

Third, the latest FTT studies, including T-ISTGNN~\cite{QiWBLYA23} and pFedCTP~\cite{Zhang00XLC24}, intend to safeguard the data privacy of traffic knowledge transfer using federated learning. Specifically, T-ISTGNN~\cite{QiWBLYA23} combines privacy-preserving traffic knowledge transfer with inductive spatio-temporal GNNs for cross-region traffic prediction. Next, pFedCTP~\cite{Zhang00XLC24} employs personalized FL to decouple the ST-Net into shared and private components, addressing spatial and temporal heterogeneity for improved local model personalization. However, both T-ISTGNN~\cite{QiWBLYA23} and pFedCTP~\cite{Zhang00XLC24} face challenges such as privacy leakage risk, data distribution discrepancies, low data quality, and high knowledge transfer overhead, making them unsuitable for real-world applications, as shown in Fig~\ref{fig:intro2}.

\textbf{\textit{In this paper, we propose an effective, efficient, and privacy-aware FTT framework to address the challenges of privacy protection, effectiveness, robustness, and efficiency in traffic knowledge transfer within federated environments.}}
\section{Problem Definitions}
\label{sec:preliminary}

\begin{table}[tb]
    \centering
    \caption{Notations and Descriptions}
    \vspace{-2mm}
    \resizebox{\linewidth}{!}{\begin{tabular}{cl}
    \hline
         \textbf{Notation} & \textbf{Description} \\ \hline
         $m$, $\mathcal{M}$ &A sensor and a set of sensors $\{m_1, m_2,\ldots\}$ \\ 
         $\mathcal{E}, A$ &A set of edges and the weighted adjacent matrix of edges  \\ 
         $\mathcal{G}$ &A road network $(\mathcal{M}, \mathcal{E}, A)$ \\ 
         $t$, $r$ &The time and training round  \\ 
         $M_t$ &A set of available sensors $\{m_i|i\leq |\mathcal{M}|\}$ at time \(t\)  \\ 
         $X_t$, $\mathcal{X}$ &The traffic data at time \(t\)  and a set of traffic data $\{X_1,X_2,\ldots\}$\\ 
         $D$ &A traffic dataset $\{X_1,X_2,\ldots;\mathcal{G}\}$\\ 
         $c$, $s$ &A client and the server\\ 
         $R$, $S$ &A source city and the target city\\ 
         $n$ &The number of clients and source cities\\ 
         $\mathcal{C}$, $\mathcal{R}$ &A set of clients $\{c_1,c_2,\ldots,c_n\}$ and source cities $\{R_1,R_2,\ldots,R_n\}$\\ 
         $\theta$, $\mathcal{L}(\cdot)$ &A model and a loss function\\ 
         $v_t^i$, $V_t$ &The $i$-level traffic subview and a traffic view $\{v_t^1, v_t^2,\ldots\}$ at time \(t\)  \\ 
         $\mathcal{P}$ &A traffic domain prototype \\ 
         \hline
    \end{tabular}}
    \vspace{-2mm}
    \label{tab:notation}
\end{table}

Table~\ref{tab:notation} presents the frequently used notations.

\textbf{Definition 1 (Road Network).} \textit{The road network is a weighted graph \(\mathcal{G} = (\mathcal{M}, \mathcal{E}, A)\), where \(\mathcal{M} = \{m_1, m_2, \dots\}\) is the set of sensors, \(\mathcal{E} \subseteq \mathcal{M} \times \mathcal{M}\) is the set of edges, and \(A \in \mathbb{R}^{|\mathcal{M}| \times |\mathcal{M}|}\) is the weighted adjacency matrix of edges. Here, \(m_i\) denotes the sensor with index \(i\).}

\textbf{Definition 2 (Traffic Data).} \textit{Given the available sensors \(M_t = \{m_i \mid i \leq |\mathcal{M}|\}\), where \(n_i\) represents an available sensor at time \(t\) and \(|M_t| \leq |\mathcal{M}|\), the traffic data is denoted as \(\mathcal{X} = \{X_1, X_2, \dots\}\), where \(X_t\in \mathbb{R}^{|M_t| \times F_1}\) is the traffic data of \(|M_t|\) available sensors at time \(t\). Here, \(F_1\) denotes the number of traffic data features. For instance, \(F_1 = 3\) when the traffic data includes flow, speed, and occupancy.}



\underline{\textbf{Problem Formulation (FTT).}}
In federated learning, multiple clients $\mathcal{C}=\{c_1, c_2, \ldots, c_n \}$ collaboratively train a global model using their local data. In the first stage, FTT trains a traffic model \(\theta_{\textit{TP}}\) to learn traffic knowledge from source cities \(\mathcal{R} = \{R_1, R_2, \ldots, R_n\}\), where each source city \(R_i\) corresponds to a client \(c_i\) shown below:
\begin{equation}
\begin{aligned}
\label{eq2}
\mathop{\min}_{\theta_{\textit{TP}}} \frac{1}{n}\sum_{i=1}^{n} \mathcal{L}(\theta_{\textit{TP}}, D^{R_i}),
\end{aligned}
\end{equation}
where \(\mathcal{L(\cdot)}\) is the loss function, and \(D^{R_i}=\{X^{R_i}_1, X^{R_i}_2,\ldots;\mathcal{G}^{R_i}\}\) is the traffic dataset of the source city \(R_i\). Here, $\mathcal{G}^{R_i}$ and \(X^{R_i}_t\) are the road network and the traffic data at time $t$ of the source city \(R_i\). In the second stage, FTT predicts the next \(T'\) traffic data based on the \(T\) historical observations at time \(t\) to transfer traffic knowledge to the target city \(S\), as formally shown below:
\begin{equation}
\label{eq2.5}
\{X^S_{t-T+1},X^S_{t-T+2}, ..., X^S_{t}; \mathcal{G}^S\} \stackrel{\theta_{\textit{TP}}}{\longrightarrow} \{X^S_{t+1},X^S_{t+2}, ..., X^S_{t+T'}\},
\end{equation}
where \(D^S=\{X^S_1, X^S_2,...;\mathcal{G}^S\}\) is the traffic dataset of the target city \(S\). Here, $\mathcal{G}^S$ and \(X^S_t\) are the road network and the traffic data at time $t$ of the target city \(S\).


\section{Our Methods}
\label{sec:framework}

\begin{figure*}[t]
    \centering
    \vspace{-4mm}
    \includegraphics[width=\textwidth]{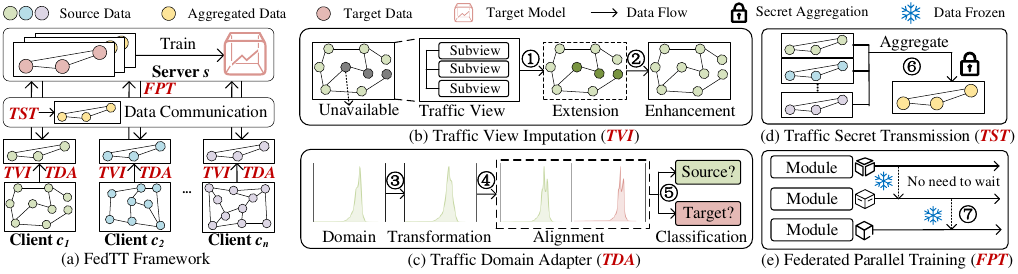}
    \vspace{-6mm}
    \caption{The Architecture of the Proposed FedTT Framework that Consists of Four Key Modules: TVI, TDA, TST, and FPT}
    \vspace{-4mm}
    \label{fig:framework}
\end{figure*}

Fig.~\ref{fig:framework} illustrates the architecture of the proposed FedTT framework, which comprises four modules: Traffic View Imputation (TVI), Traffic Domain Adapter (TDA), Traffic Secret Transmission (TST), and Federated Parallel Training (FPT). As shown in Fig.~\ref{fig:framework}(a), FedTT comprises $n$ clients $\mathcal{C}=\{c_1,c_2,\ldots,c_n\}$ and a central server $s$. Specifically, each source city \(R_i\) is treated as a client \(c_i\), while the target city \(S\) is treated as the server \(s\). The traffic domains of the data in clients are transformed to align with the server's domain, and the server's traffic model is trained on this transformed data uploaded by clients. Consequently, the FTT problem defined in Eqs.~\ref{eq2} and~\ref{eq2.5} is reformulated to minimize the sum of the following losses:
\begin{equation}
\label{eq3}
\mathop{\min}_{\theta_{\textit{TP}}} \frac{1}{n}\sum_{i=1}^{n} \mathcal{L}(\theta_{\textit{TP}}, D^{R_i\to S}, D^S),
\end{equation}
where \(D^{R_i \to S}\) represents the traffic dataset whose traffic domain is transformed from the source city \(R_i\) to the target city \(S\). 


\begin{itemize}[leftmargin=*]
\item{}
\textbf{TVI Module.} The TVI module is designed to improve the quality of traffic data, enabling the framework to achieve robust training performance even with data of average quality, as shown in Fig.~\ref{fig:framework}(b). Before initiating the training process in the FedTT framework, the module first analyzes spatial dependencies within the traffic data to extend the traffic view (\ding{172}). It then captures temporal dependencies to enhance the extended traffic view (\ding{173}).

\item{}
\textbf{TDA Module.} To address discrepancies in traffic data distribution between source and target cities, the TDA module adapts data of source cities to align with the target city's traffic domain, as shown in Fig.~\ref{fig:framework}(c). During the FedTT framework training process, it conducts traffic domain transformation and alignment for the source cities' data (\ding{174}--\ding{175}). Then, the module performs traffic domain classification to categorize the traffic data domain (\ding{176}).

\item{}
\textbf{TST Module.} To ensure data privacy during data aggregation, the TST module employs the proposed traffic secret aggregation method to securely transmit and aggregate the transformed data from source cities (\ding{177}), as shown in Fig.~\ref{fig:framework}(d).

\item{}
\textbf{FPT Module.} To enhance the training efficiency of FedTT, the FPT module leverages split learning and parallel optimization techniques to decompose the training process and freeze the data required by both the client and server (\ding{178}), as shown in Fig.~\ref{fig:framework}(e).

\end{itemize}

\subsection{Traffic View Imputation}
\begin{figure}[t]
    \centering
    \includegraphics[width=0.47\textwidth]{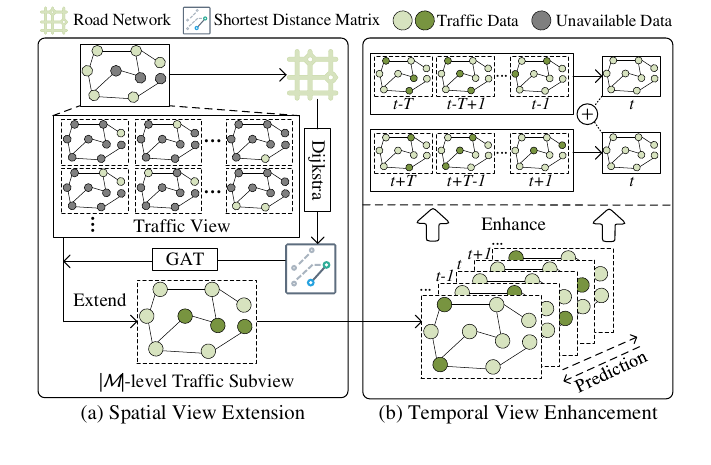}
    \vspace{-4mm}
    \caption{The Process of Traffic View Imputation}
    \vspace{-6mm}
    \label{fig:tvi}
\end{figure}

\textbf{Design Motivation.} Existing federated traffic transfer methods often overlook the challenges associated with low-quality traffic data, especially when missing data is prevalent, thereby significantly undermining the performance of traffic knowledge transfer models. Drawing inspiration from data imputation techniques~\cite{hty24,xyc22,wcp23} for addressing missing data, we propose the Traffic View Imputation (TVI) module, as shown in Fig.~\ref{fig:tvi}. This module enhances traffic data quality by completing and predicting missing traffic data through a comprehensive exploration of the spatial and temporal dependencies inherent in traffic data, as formally shown below:
\begin{equation}
\label{eq4}
    \{X_1,X_2,\ldots ; \mathcal{G} \} \xrightarrow{\theta_{\textit{TVI}}} \{\widetilde{X}_1,\widetilde{X}_2,\ldots \}, X_t \in \mathbb{R}^{|M_t| \times F_1},
\end{equation}
where $\theta_{\textit{TVI}}$ is the traffic view imputation module consisting of a spatial view extension model $\theta_{\textit{SV}}$ and a temporal view extension model $\theta_{\textit{TV}}$. Besides, $\widetilde{X}_t \in \mathbb{R}^{|\mathcal{M}| \times F_1}$ is the predicted traffic data of $|\mathcal{M}|$ sensors. In addition, the traffic view represents traffic data of all sensors at a certain time, as defined below.


\textbf{Definition 4 (Traffic View).} \textit{A traffic view is the snapshot of traffic data of sensors $\mathcal{M}$ at time \(t\), consisting of a set of multi-level traffic subviews, denoted as $V_t = \{v_t^1, v_t^2,\ldots v_t^{|M_t|}\}$, where $i$-level traffic subview $v_t^i$ is a set of traffic data of $i$ sensors at time \(t\).}

\textbf{i) Spatial View Extension.}
In the first stage, TVI extends the $|\mathcal{M}|$-level traffic subview at time \(t\), addressing the first problem to be solved in Eq.~\ref{eq4}, as formally outlined below:
\begin{equation}
\label{eq:6}
    \{v_t^1, v_t^2,\ldots v_t^{|M_t|};\mathcal{G}\} \xrightarrow{\theta_{\textit{SV}}} \textit{sv}_t^{|\mathcal{M}|},
\end{equation}
Where $\theta_{\textit{SV}}$ denotes the spatial view extension model, and $\textit{sv}_t^{|\mathcal{M}|}$ represents the $|\mathcal{M}|$-level traffic subview at time $t$ in the spatial view extension. As shown in Fig.~\ref{fig:tvi}(a), it first computes the shortest distance matrix $\mathcal{A}=\{A_1, A_2, \ldots , A_{|\mathcal{M}|}\}$, where $A_i$ represents the shortest distance tensor of sensor $m_i$ to other sensors. This is computed using Dijkstra's algorithm~\cite{ewd22} with the weighted adjacency matrix $A$. Next, the feature of each sensor is computed:
\begin{equation}
h_i=\theta_{\textit{GAT}}(A_i), i=1,2,\ldots ,|\mathcal{M}|,
\end{equation}
where $h_i \in \mathbb{R}^{K \times F_2}$ represents the $K$-head feature of sensor $m_i$ with $F_2$ feature dimensions, and $\theta_{\textit{GAT}}$ is the Graph Attention Network (GAT) model~\cite{Velickovic18} with $K=8$ and $F_2=128$. Additionally, the extension result of multi-level traffic subviews from the traffic view $V_t$ is averaged to obtain the $|\mathcal{M}|$-level traffic subview:
\begin{equation}
   \textit{sv}_t^{|\mathcal{M}|}=\frac{1}{|V_t|} \sum^{|V_t|}_{i=1} \frac{1}{|v_t^i|} \sum^{|v_t^i|}_{j=1} \theta_E(\frac{1}{i}\sum^{i}_{k=1}(H(v_t^i[j][k])\cdot{(v_t^i[j][k])}^\top)),
\end{equation}
where $v_t^i[j][k]$ represents the traffic data of the $k$-th sensor in the $j$-th combination within the $i$-level traffic subview at time $t$, ${(v_t^i[j][k])}^\top \in \mathbb{R}^{1 \times F_1}$ denotes the transpose matrix of $v_t^i[j][k]$, and $H(v_t^i[j][k]) \in \mathbb{R}^{K \times F_2 \times 1}$ represents the multi-head feature of the sensor corresponding to $v_t^i[j][k]$. The term $H(v_t^i[j][k]) \cdot {(v_t^i[j][k])}^\top \in \mathbb{R}^{K \times F_2 \times F_1}$ is the matrix multiplication of $H(v_t^i[j][k])$ and ${(v_t^i[j][k])}^\top$. Additionally, $\theta_E$ is a fully connected layer with $K \times F_2 \times F_1$ input dimensions and $|\mathcal{M}| \times F_1$ output dimensions.

Finally, it uses the traffic data from available sensors as the ground truth and computes the loss between the predicted traffic data of these sensors and the ground truth to train the $\theta_{\textit{SV}}$ model, which includes $\theta_{\textit{GAT}}$ and $\theta_{\textit{E}}$, as shown below:
\begin{equation}
\label{eq8}
\mathop{\min}_{\theta_{\textit{SV}}} \mathcal{L}({\theta_{\textit{SV}}}, \mathcal{V}_\textit{SV})=\mathop{\min}_{\theta_{\textit{SV}}} \frac{1}{|\mathcal{V}_\textit{SV}|}\sum^{|\mathcal{V}_\textit{SV}|}_{t=1} \frac{1}{|M_t|}(\textit{sv}_t^{|M_t|}-X_t),
\end{equation}
where $\mathcal{V}_\textit{SV}=\{\textit{sv}_1^{|\mathcal{M}|}, \textit{sv}_2^{|\mathcal{M}|},\ldots \}$ is the set of traffic subviews at different times obtained by spatial view extension, and $\textit{sv}_t^{|M_t|}$ is the predicted traffic data of available sensors at time $t$. In this way, TVI gets the extended $|\mathcal{M}|$-level traffic subview $\textit{sv}_t^{|\mathcal{M}|}$ with the traffic data of available sensors $X_t$ and the predicted traffic data of unavailable sensors $\textit{sv}_t^{|\mathcal{M}-M_t|}$, as shown below:
\begin{equation}
\textit{sv}_t^{|\mathcal{M}|}=(X_t, \textit{sv}_t^{|\mathcal{M}-M_t|})
\end{equation}

\textbf{ii) Temporal View Enhancement.} 
In the second stage, TVI predicts the next/previous $T'$ $|\mathcal{M}|$-level traffic subviews based on the preceding/succeeding $T$ $|\mathcal{M}|$-level traffic subviews and averages them, thereby enhancing the $|\mathcal{M}|$-level traffic subview, as shown in Fig.~\ref{fig:tvi}(b). This represents the second problem that must be addressed in Eq.~\ref{eq4}, as formally demonstrated below:
\begin{equation}
\label{eq10}
\begin{aligned}
   \{\textit{sv}_{t-T}^{|\mathcal{M}|},&\textit{sv}_{t-T+1}^{|\mathcal{M}|},\ldots ,\textit{sv}_{t-1}^{|\mathcal{M}|}\} \xrightarrow{\theta_{\textit{TV}}} \textit{ntv}_{t}^{|\mathcal{M}|},\\
   \{\textit{sv}_{t+T}^{|\mathcal{M}|},&\textit{sv}_{t+T-1}^{|\mathcal{M}|},\ldots ,\textit{sv}_{t+1}^{|\mathcal{M}|}\} \xrightarrow{\theta_{\textit{TV}}}  \textit{ptv}_{t}^{|\mathcal{M}|},\\
   &\textit{tv}_{t}^{|\mathcal{M}|}= \frac{1}{2}(\textit{ntv}_{t}^{|\mathcal{M}|}+ \textit{ptv}_{t}^{|\mathcal{M}|}),
\end{aligned}
\end{equation}
where $T' = 1$ since short-term traffic prediction generally outperforms long-term traffic prediction in most traffic prediction studies~\cite{zys23,jjh23,jjw23}. Besides, $\textit{ntv}_t^{|\mathcal{M}|}$ denotes the next $|\mathcal{M}|$-level traffic subview, $\textit{ptv}_t^{|\mathcal{M}|}$ denotes the previous $|\mathcal{M}|$-level traffic subview, and $\textit{tv}_t^{|\mathcal{M}|}$ represents the $|\mathcal{M}|$-level traffic subview at time $t$ in the temporal view enhancement. Additionally, $\theta_{\textit{TV}}$ is the temporal view enhancement model, which employs the state-of-the-art DyHSL traffic model~\cite{zys23} for temporal view enhancement. It then uses the traffic data from available sensors as the ground truth and computes the loss between the predicted traffic data of the available sensors and the ground truth to train the $\theta_{\textit{TV}}$ model, as shown below:
\begin{equation}
\label{eq12}
\mathop{\min}_{\theta_{\textit{TV}}} \mathcal{L}({\theta_{\textit{TV}}}, V^{|\mathcal{M}|})=\mathop{\min}_{\theta_{\textit{TV}}} \frac{1}{|V^{|\mathcal{M}|}|}\sum^{|V^{|\mathcal{M}|}|}_{t=1} \frac{1}{|M_t|}(\textit{tv}_t^{|M_t|}-X_t),
\end{equation}
where $\mathcal{V}_\textit{TV}=\{\textit{tv}_1^{|\mathcal{M}|}, \textit{tv}_2^{|\mathcal{M}|},\ldots \}$ represents the set of traffic subviews at different times obtained by temporal view enhancement, and $\textit{tv}_t^{|M_t|}$ is the predicted traffic data of the available sensors at time $t$. In this manner, TVI enhances the $|\mathcal{M}|$-level traffic subview $\textit{tv}_t^{|\mathcal{M}|}$ by combining the traffic data from available sensors $X_t$ and the predicted traffic data from unavailable sensors $\textit{tv}_t^{|\mathcal{M}-M_t|}$:
\begin{equation}
\widetilde{X}_t=\textit{tv}_t^{|\mathcal{M}|}=(X_t, \textit{tv}_t^{|\mathcal{M}-M_t|})
\end{equation}

Note that the training of the TVI model should be completed before the training of the FedTT framework, as it only needs to be conducted within each city.

\subsection{Traffic Domain Adapter}
\begin{figure}[t]
    \centering
    \includegraphics[width=0.49\textwidth]{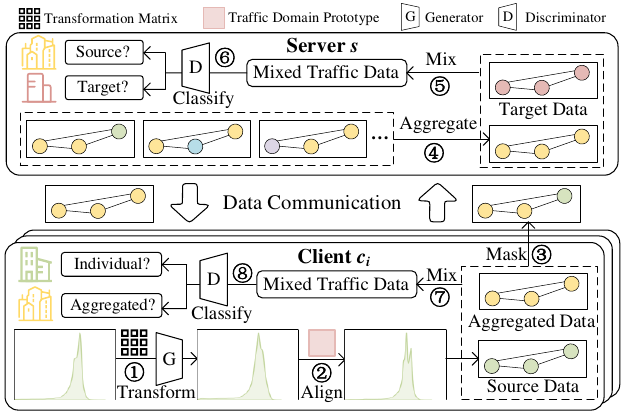}
    \vspace{-6mm}
    \caption{TDA and TST modules}
    \vspace{-6mm}
    \label{fig:tda&tst}
\end{figure}

\textbf{Design Motivation.} None of the existing approaches consider traffic data distribution discrepancies between the source and target cities when performing cross-city knowledge transfer, which decreases the effectiveness of traffic knowledge transfer. 
Motivated by this, to reduce the impact of traffic data distribution discrepancies on model performance, we propose the Traffic Domain Adapter (TDA) module, as shown in Fig.~\ref{fig:tda&tst}. This module reduces traffic domain discrepancies by uniformly transforming data from the traffic domain of the source city ("source domain" for short) to the traffic domain of the target city ("target domain" for short) shown below:
\begin{equation}
\{\widetilde{X}^R_1,\widetilde{X}^R_2,\ldots \} \xrightarrow{\theta_{\textit{TDA}}} \{X^{R\to S}_1,X^{R\to S}_2,\ldots \},
\end{equation}
where $\widetilde{X}^R_t$ is the predicted traffic data of $|\mathcal{M}^R|$ sensors obtained by the TVI module, $X_t^{R \to S}$ is the transformed data of $|\mathcal{M}^S|$ sensors, and $\theta_{\textit{TDA}}$ is a generative adversarial network~\cite{Wanghw18} consisting of a generator model $\theta_{\textit{Gen}}$ and a discriminator model $\theta_{\textit{Dis}}$.

\textbf{i) Traffic Domain Transformation.}
In the first step, TDA uses the generator model, road network, and traffic domain prototype to transform the traffic data from the source domain to the target domain, as shown in Fig.~\ref{fig:tda&tst} (\ding{172}), where the traffic domain prototype is the representative traffic sample that can reflect the main feature of traffic data in the domain, as defined below.

\textbf{Definition 5 (Traffic Domain Prototype).} \textit{Given the traffic data $\mathcal{X}=\{X_1, X_2,\ldots \}$ in a traffic domain, a traffic domain prototype $\mathcal{P}$ is the central traffic data, which is computed as the averaged value of all traffic data, as formally shown below:}
\begin{equation}
\mathcal{P}=\frac{1}{|\mathcal{X}|} \sum^{|\mathcal{X}|}_{t=1}X_t
\end{equation}

First, it computes the transformation matrix $A_\mathcal{G}$ of the road network, as shown below:
\begin{equation}
{(A_{\mathcal{G}})}^\top \cdot \mathcal{G}^R \cdot A_{\mathcal{G}}= \mathcal{G}^S,
\end{equation}
where $A_{\mathcal{G}}$ can learn the road network information of the source and target cities. Here, $A_{\mathcal{G}} \in \mathbb{R}^{|\mathcal{M}^S| \times |\mathcal{M}^R|}$ is computed by the gradient descent method~\cite{rh51}, where $|\mathcal{M}^R$|and $|\mathcal{M}^S$| denote the number of sensors in the source city $R$ and the target city $S$, respectively. Similarly, it then computes the transformation matrix $A_\mathcal{P}$ of the traffic domain prototype, as shown below:
\begin{equation}
A_\mathcal{P} \cdot \mathcal{P}^R = \mathcal{P}^S,
\end{equation}
where $\mathcal{P}^R$ and $\mathcal{P}^S$ are traffic domain prototypes of the source and target cities, respectively. $A_\mathcal{P} \in \mathbb{R}^{|\mathcal{M}^S| \times |\mathcal{M}^R|}$ can learn the traffic domain prototype information of the source and target cities, which is computed by the gradient descent method. Then, the generator model transforms the traffic data from the source domain to the target domain using the above transformation matrices $A_\mathcal{G}$ and $A_\mathcal{P}$, as shown below:
\begin{equation}
X^{R\to S}_t = \theta_{\mathcal{G}}(A_{\mathcal{G}} \cdot \widetilde{X}^R_t)+\theta_{\mathcal{P}}(A_\mathcal{P} \cdot \widetilde{X}^R_t)+\theta_X(\widetilde{X}^R_t),
\end{equation}
where $\theta_{\mathcal{G}}$ and $\theta_{\mathcal{P}}$ are both a three-layer MLP (i.e., Multi-Layer Perception) model~\cite{rf58} with $|\mathcal{M}^S| \times F_2$ input dimensions, $1024$ hidden dimensions, and $|\mathcal{M}^S| \times F_1$ output dimensions. Besides, $\theta_X$ is a three-layer MLP model with $|\mathcal{M}^R| \times F_1$ input dimensions, $1024$ hidden dimensions, and $|\mathcal{M}^S| \times F_1$ output dimensions. Here, $\mathcal{M}^R$ and $\mathcal{M}^S$ are sensors of the source and target cities, respectively.

\textbf{ii) Traffic Domain Alignment.}
In the second step, TDA trains the generator model $\theta_\textit{Gen}$, which consists of $\theta_{\mathcal{G}}$, $\theta_{\mathcal{P}}$, and $\theta_X$, as shown in Fig.~\ref{fig:tda&tst} (\ding{173}). Specifically, it aligns the transformed data $\mathcal{X}^{R\to S}=\{X_1^{R\to S}, X_2^{R\to S},\ldots \}$ of the source city with the traffic domain prototype $\mathcal{P}^S$ of the target city $S$, as described below:
\begin{equation}
\label{eq19}
\mathop{\min}_{\theta_\textit{Gen}} \mathcal{L}({\theta_\textit{Gen}}, \mathcal{X}^{R\to S})=\mathop{\min}_{\theta_\textit{Gen}}\frac{1}{|\mathcal{X}^{R\to S}|}\sum^{|\mathcal{X}^{R\to S}|}_{t=1} \frac{1}{|\mathcal{M}^S|}(X_t^{R\to S}-\mathcal{P}^S),
\end{equation}

\textbf{iii) Traffic Domain Classification.}
In the third step, TDA trains the discriminator model $\theta_{\textit{Dis}}$ to classify the traffic data domain (\ding{176}--\ding{177} shown in Fig.~\ref{fig:tda&tst}), as shown below:
\begin{equation}
\theta_{\textit{Dis}}(X_t^{\textit{RS}} \in \mathcal{X}^{\textit{RS}})= \begin{cases}
\begin{aligned}
&P(X_t^{\textit{RS}} \in \mathcal{X}^{R \to S}) \\
&P(X_t^{\textit{RS}} \in \mathcal{X}^S)
\end{aligned}
\end{cases},
\end{equation}
where $\mathcal{X}^{\textit{RS}}=\{X_1^{\textit{RS}}, X_2^{\textit{RS}},\ldots \}$ is the traffic data mixed with the transformed data $\mathcal{X}^{R\to S}$ of the source city and the traffic data $\mathcal{X}^S$ of the target city. Besides, $P(X_t^{\textit{RS}} \in \mathcal{X}^{R \to S})$ and $P(X_t^{\textit{RS}} \in \mathcal{X}^S)$ are the possibility of $X_t^{\textit{RS}} \in \mathcal{X}^{R \to S}$ and $X_t^{\textit{RS}} \in \mathcal{X}^S$, respectively. Additionally, $\theta_{\textit{Dis}}$ is a three-layer MLP model with $|\mathcal{M}^S| \times F_1$ input dimensions, 1024 hidden dimensions, and 2 output dimensions. Then, the training process of the discriminator model $\theta_{\textit{Dis}}$ is shown:
\begin{equation}
\label{eq21}
\begin{aligned}
&\mathop{\min}_{\theta_{\textit{Dis}}} \mathcal{L}({\theta_{\textit{Dis}}}, \mathcal{X}^{\textit{RS}})=\\ 
&\mathop{\min}_{\theta_{\textit{Dis}}}\frac{1}{|\mathcal{X}^{\textit{RS}}|}\sum^{|\mathcal{X}^{\textit{RS}}|}_{t=1} 
\begin{cases}
\begin{aligned}
&-log(P(X_t^{\textit{RS}} \in \mathcal{X}^{R \to S})), \ \ if \ X_t^{\textit{RS}} \in \mathcal{X}^{R \to S}\\
&-log(P(X_t^{\textit{RS}} \in \mathcal{X}^S))\ \ \ \ \ , \ \ if \ X_t^{\textit{RS}} \in \mathcal{X}^S
\end{aligned}
\end{cases}
\end{aligned}
\end{equation}
Next, we update the training process of the generator model $\theta_{\textit{Gen}}$ in Eq.~\ref{eq19}, as shown below:
\begin{equation}
\small
\label{eq22}
\begin{aligned}
\mathop{\min}_{\theta_{\textit{Gen}}} \mathcal{L}({\theta_{\textit{Gen}}}, \theta_{\textit{Dis}}, \mathcal{X}^{R\to S}, \mathcal{X}^{\textit{RS}})=\mathop{\min}_{\theta_{\textit{Gen}}} \mathcal{L}({\theta_{\textit{Gen}}}, \mathcal{X}^{R\to S})-\lambda_1 \mathcal{L}(\theta_{\textit{Dis}}, \mathcal{X}^{\textit{RS}}),
\end{aligned}
\end{equation}
where $\lambda_1$ is the hyper-parameter to control the trade-off between generator loss and discriminator loss, which is determined by the Parameter Sensitivity~\ref{sec:para}.

\subsection{Traffic Secret Transmission}
\textbf{Design Motivation.} 
Existing FTT works upload training gradients or model parameters for model aggregation in the federated training, where attackers derive the traffic data from the uploaded model parameters or training gradients through inference attacks~\cite{kg24,ybw24,llz24}. Although some techniques such as Homomorphic Encryption (HE)~\cite{rlr78} and Differential Privacy (DP)~\cite{cd06} can be employed for federated secure aggregation, HE introduces significant computational and communication overheads, reducing training efficiency, while DP reduces the data utility, leading to lower model accuracy. In contrast, to protect traffic data privacy, we design the Traffic Secret Transmission (TST) module to securely transmit and aggregate the transformed data from source cities by a lightweight traffic secret aggregation method without sacrificing the training efficiency or model accuracy, as shown in Fig.~\ref{fig:tda&tst} (\ding{174}--\ding{175}).

First, it aggregates the transformed data from source cites:
\vspace{-2mm}
\begin{equation}
\label{eq23}
\overline{\mathcal{X}}^{\mathcal{R}\to S}_{(r)} = \frac{1}{n} \sum^{n}_{i=1} \mathcal{X}^{R_i\to S}_{(r)},
\end{equation}
\vspace{-2mm}

\noindent where $\mathcal{X}^{R_i\to S}_{(r)}$ is the transformed data from the source city $R_i$ at round $r$ , and $\overline{\mathcal{X}}^{\mathcal{R}\to S}_{(r)}$ is the aggregated data at round $r$. In this way, it not only protects traffic data privacy as information for individual source cities can not be inferred or derived from the aggregated data~\cite{phw24,wd24} but also reduces the influence of specific traffic patterns of source cities on traffic models when the target city uses the aggregated data for model training.

However, since the above data aggregation requires transmitting transformed data from all clients to the server, it exposes individual transformed data, thereby failing to provide effective privacy protection. To protect individual transformed data during the aggregation process, it designs a privacy-preserving and lightweight traffic secret aggregation method without compromising the training efficiency or model accuracy. Specifically, it first masks the transformed data based on the aggregated data, as shown below: 
\begin{equation}
\label{eq24}
\mathcal{X}_{(r)}^{(\mathcal{R}\to S,\ R_i)}=\overline{\mathcal{X}}_{(r-1)}^{\mathcal{R}\to S}+\mathcal{X}^{R_i\to S}_{(r)}-\mathcal{X}^{R_i\to S}_{(r-1)},
\end{equation}
where $\mathcal{X}^{(\mathcal{R}\to S,\ R_i)}_{(r)}$ is the mask data computed in the source city $R_i$ (i.e., the client $c_i$) and transmitted to the server at round $r$. Note that, when $r=0$, the client needs to transmit its transformed data to other clients for initial aggregation. Then, the server computes the sum of mask data from all source cities at round $r$:
\begin{equation}
\begin{aligned}
\sum^{n}_{i=1} \mathcal{X}_{(r)}^{(\mathcal{R}\to S,\ R_i)}&=n*\overline{\mathcal{X}}_{(r-1)}^{\mathcal{R}\to S}+\sum^{n}_{i=1}\mathcal{X}^{R_i\to S}_{(r)}-\sum^{n}_{i=1}\mathcal{X}^{R_i\to S}_{(r-1)}\\
&=n*\overline{\mathcal{X}}_{(r-1)}^{\mathcal{R}\to S}+\overline{\mathcal{X}}_{(r)}^{\mathcal{R}\to S}-\overline{\mathcal{X}}_{(r-1)}^{\mathcal{R}\to S}\\
&=(n-1)*\overline{\mathcal{X}}_{(r-1)}^{\mathcal{R}\to S}+\overline{\mathcal{X}}_{(r)}^{\mathcal{R}\to S}
\end{aligned}
\end{equation}
Finally, the server gets the aggregated data at round $r$  based on the aggregated data $\overline{\mathcal{X}}_{(r-1)}^{\mathcal{R}\to S}$ at round $r-1$, as shown below:
\begin{equation}
\label{eq25}
\overline{\mathcal{X}}_{(r)}^{\mathcal{R}\to S}=\sum^{n}_{i=1} \mathcal{X}_{(r)}^{(\mathcal{R}\to S,\ R_i)}-(n-1)*\overline{\mathcal{X}}_{(r-1)}^{\mathcal{R}\to S}
\end{equation}
In this way, traffic secret aggregation ensures that the server can only access the aggregated data without revealing the individual transformed data. Since the data transmission is the aggregated data instead of individual transformed data, the client $c_i$ can train a discriminator model $\theta_{\textit{Dis}}^{R_i}$ in the source city $R_i$ to classify the aggregated data and individual transformed data to improve the generator model performance in TDA (\ding{178}--\ding{179} shown in Fig.~\ref{fig:tda&tst}): 
\begin{equation}
\theta_{\textit{Dis}}^{R_i}(X^{R_i S}_t \in \mathcal{X}^{R_i S})= \begin{cases}
\begin{aligned}
&P(X^{R_i S}_t \in \mathcal{X}^{R_i \to S}) \\
&P(X^{R_i S}_t \in \overline{\mathcal{X}}^{\mathcal{R} \to S})
\end{aligned}
\end{cases},
\end{equation}
where $\mathcal{X}^{R_i S}=\{X^{{R_i S}}_1, X^{{R_i S}}_2,\ldots \}$ is the traffic data mixed with the aggregated data $\overline{\mathcal{X}}^{\mathcal{R} \to S}$ and transformed data $\mathcal{X}^{R_i \to S}$ of the source city $R_i$. Besides, $\theta_{\textit{Dis}}^{R_i}$ is a three-layer MLP model with $|\mathcal{M}^S| \times F_1$ input dimensions, 1024 hidden dimensions, and 2 output dimensions. Then, the training process of the discriminator model $\theta_{\textit{Dis}}^{R_i}$ is shown:
\begin{equation}
\label{eq28}
\begin{aligned}
&\mathop{\min}_{\theta_{\textit{Dis}}^{R_i}} \mathcal{L}({\theta_{\textit{Dis}}^{R_i}}, \mathcal{X}^{R_i S})=\\ 
&\mathop{\min}_{\theta_{\textit{Dis}}^{R_i}}\frac{1}{|\mathcal{X}^{R_i S}|} \sum^{|\mathcal{X}^{R_i S}|}_{t=1} 
\begin{cases}
\begin{aligned}
&-log(P(X^{R_i S}_t\in \mathcal{X}^{R_i \to S})), \ \ if \ X^{R_i S}_t\in \mathcal{X}^{R_i \to S}\\
&-log(P(X^{R_i S}_t \in \mathcal{X}^{\mathcal{R} \to S})), \ \ if \ X^{R_i S}_t \in \mathcal{X}^{\mathcal{R} \to S}
\end{aligned}
\end{cases}
\end{aligned}
\end{equation}
Therefore, given the traffic data $\mathcal{X}^{\mathcal{R}S}=\{X^{\mathcal{R}S}_1, X^{\mathcal{R}S}_2,\ldots \}$ consisting of aggregated data $\overline{\mathcal{X}}^{\mathcal{R} \to S}$ and traffic data $\mathcal{X}^S$ of the target city $S$, the training process of the generator model $\theta_{\textit{Gen}}$ in Eq.~\ref{eq22} is updated, as shown below:
\begin{equation}
\begin{aligned}
&\mathop{\min}_{\theta_{\textit{Gen}}^{R_i}} \mathcal{L}(\theta_{\textit{Gen}}^{R_i}, \theta_{\textit{Dis}}, \theta_{\textit{Dis}}^{R_i}, \mathcal{X}^{R_i\to S}, \mathcal{X}^{\mathcal{R}S}, \overline{\mathcal{X}}^{R_i S})=
\\ &\mathop{\min}_{\theta_{\textit{Gen}}^{R_i}} \mathcal{L}(\theta_{\textit{Gen}}^{R_i}, \mathcal{X}^{R_i\to S})- \lambda_1 \mathcal{L}(\theta_{\textit{Dis}}, \mathcal{X}^{\mathcal{R}S}) - \lambda_2 \mathcal{L}(\theta_{\textit{Dis}}^{R_i}, \mathcal{X}^{R_i S}),
\label{eq29}
\end{aligned}
\end{equation}
where $i=1,2,\ldots,{n}$, $\theta_{\textit{Gen}}^{R_i}$ is the generator model of the source city $R_i$ (i.e., the client $c_i$), $\theta_{\textit{Dis}}$ is the discriminator model of the target city $S$ (i.e., the server $s$), and $\theta_{\textit{Dis}}^{R_i}$ is the discriminator model of the source city $R_i$. Besides, $\lambda_1$ and $\lambda_2$ are the hyper-parameter to control the trade-off between generator loss and discriminator loss, which is determined by the Parameter Sensitivity~\ref{sec:para}.


\subsection{Federated Parallel Training}
\textbf{Design Motivation.} 
The traffic knowledge transfer process in existing FTT methods suffers from inefficiency. Although we propose the FedTT framework, which transforms the traffic domain of data and trains the model on transformed data, it remains certain computation and communication overheads. Specifically, the sequential training of several models, including the generator model, discriminator model, and traffic model, hampers the overall training efficiency of the proposed framework. Besides, the transmission of transformed data and the backpropagation of model gradients still introduce huge communication costs. To improve the training efficiency, we introduce the Federated Parallel Training (FPT) module to reduce the data transmission and train the models in parallel.


\textbf{i) Split Learning.} 
To reduce the communication overhead and improve the training efficiency, it employs split learning~\cite{czm21} to decompose the sequential training process into the client and server training and freeze the data required by the client and server. Specifically, in the client training, the client $c_i$ is responsible for training the generator model $\theta_{\textit{Gen}}^{R_i}$ and the discriminator model $\theta_{\textit{Dis}}^{R_i}$ of the source city $R_i$. Then, it stores and freezes the classifier result computed by the server's discriminator model for training the generator model $\theta_{\textit{Gen}}^{R_i}$ in Eq.~\ref{eq29}, as shown below:
\begin{equation}
\label{eq30}
\mathop{\min}_{\theta_{\textit{Gen}}^{R_i}} \mathcal{L}(\theta_{\textit{Gen}}^{R_i}, \mathcal{X}^{R_i\to S})- \lambda_1 *\textit{Fr}(\mathcal{L}(\theta_{\textit{Dis}}, \mathcal{X}^{\mathcal{R}S})) - \lambda_2 \mathcal{L}(\theta_{\textit{Dis}}^{R_i}, \mathcal{X}^{R_i S}),
\end{equation}
where $\textit{Fr}(\cdot)$ is the frozen function and uses the historical stored data, which updates every 5 rounds. Next, it also stores and freezes the aggregated data sent by the server for training the discriminator model $\theta_{\textit{Dis}}^{R_i}$  in Eq.~\ref{eq28}, as shown below:
\begin{equation}
\label{eq31}
\small
\mathop{\min}_{\theta_{\textit{Dis}}^{R_i}}\frac{1}{|\mathcal{X}^{R_i S}|} \sum^{|\mathcal{X}^{R_i S}|}_{t=1} 
\begin{cases}
\begin{aligned}
&-log(P(X^{R_i S}_t\in \mathcal{X}^{R_i \to S})), if \ X^{R_i S}_t\in \mathcal{X}^{R_i \to S}\\
&-log(P(X^{R_i S}_t \in \mathcal{X}^{\mathcal{R} \to S})), if \ X^{R_i S}_t \in \textit{Fr}(\mathcal{X}^{\mathcal{R} \to S})
\end{aligned}
\end{cases}
\end{equation}
In the server training, the server $s$ manages the training of the discriminator model $\theta_{\textit{Dis}}$ and traffic model $\theta_{\textit{TP}}$ of the target city $S$. Then, it stores and freezes the mask data uploaded by the client to compute the aggregated data for training the discriminator model $\theta_{\textit{Dis}}$ and traffic model $\theta_{\textit{TP}}$ in Eqs.~\ref{eq21} and~\ref{eq3}, as shown below:
\begin{equation}
\small
\mathop{\min}_{\theta_{\textit{Dis}}}\frac{1}{|\mathcal{X}^{\mathcal{R}S}|}\sum^{|\mathcal{X}^{\mathcal{R}S}|}_{t=1} 
\begin{cases}
\begin{aligned}
&-log(P(X_t^{\mathcal{R}S} \in \mathcal{X}^{\mathcal{R} \to S})), \ \ if \ X_t^{\mathcal{R}S} \in \textit{Fr}(\mathcal{X}^{\mathcal{R} \to S})\\
&-log(P(X_t^{\mathcal{R}S} \in \mathcal{X}^S))\ \ \ \ \ , \ \ if \ X_t^{\mathcal{R}S} \in \mathcal{X}^S
\end{aligned}
\end{cases},
\end{equation}
\begin{equation}
\small
\mathop{\min}_{\theta_{\textit{TP}}} \mathcal{L}(\theta_{\textit{TP}}, \textit{Fr}(D^{\mathcal{R}\to S}), D^S),
\end{equation}
where $D^{\mathcal{R}\to S}$ is the aggregated dataset from source cities, In this way, it enables the simultaneous training of the client and server to reduce data transmission and improve the training efficiency. 

\textbf{ii) Parallel Optimization.} 
To further improve the training parallelism, it proposes parallel optimization to reduce data dependencies on the client and server. Specifically, the client $c_i$ caches and freezes the required local data for parallel training the generator model $\theta_{\textit{Gen}}^{R_i}$ and the discriminator model $\theta_{\textit{Dis}}^{R_i}$ of the source city $R_i$ in Eqs~\ref{eq30} and~\ref{eq31}, as shown below:
\begin{equation}
\small
\mathop{\min}_{\theta_{\textit{Gen}}^{R_i}} \mathcal{L}(\theta_{\textit{Gen}}^{R_i}, \mathcal{X}^{R_i\to S})- \lambda_1 *\textit{Fr}(\mathcal{L}(\theta_{\textit{Dis}}, \mathcal{X}^{\mathcal{R}S})) - \lambda_2*\textit{Fr}^{'}(\mathcal{L}(\theta_{\textit{Dis}}^{R_i}, \mathcal{X}^{R_i S})),
\end{equation}
\begin{equation}
\small
\mathop{\min}_{\theta_{\textit{Dis}}^{R_i}}\frac{1}{|\mathcal{X}^{R_i S}|} \sum^{|\mathcal{X}^{R_i S}|}_{t=1} 
\begin{cases}
\begin{aligned}
&-log(P(X^{R_i S}_t\in \mathcal{X}^{R_i \to S})), if \ X^{R_i S}_t\in \textit{Fr}^{'}(\mathcal{X}^{R_i \to S})\\
&-log(P(X^{R_i S}_t \in \mathcal{X}^{\mathcal{R} \to S}))\ , if \ X^{R_i S}_t \in \textit{Fr}(\mathcal{X}^{\mathcal{R} \to S})
\end{aligned}
\end{cases},
\end{equation}
where $\textit{Fr}^{'}(\cdot)$ is the frozen function and uses the historical cached data, which updates each round. Then, the server trains the discriminator model $\theta_{\textit{Dis}}$ and traffic model $\theta_{\textit{TP}}$ in parallel without data dependencies. Note that the training of the TVI model is completed before the training of the FedTT framework.

\begin{algorithm}[t]
\caption{The training of the FedTT framework in the client $c_i$}
\label{algo:1}
\KwIn{Server $s$ (target city $S$); local data $\mathcal{X}^{R_i}$; models $\theta_{\text{TVI}}, \theta_{\text{Gen}}^{R_i}, \theta_{\text{Dis}}^{R_i}$}
$\widetilde{\mathcal{X}}^{R_i} \gets \textsc{Complete}(\theta_{\text{TVI}}, \mathcal{X}^{R_i})$\tcp*{Impute missing data}
\For{$tr \gets 1$ \KwTo $T$}{
  \For{each $X^{R_i}_{(r)} \in \widetilde{\mathcal{X}}^{R_i}$}{
    $X^{R_i \to S}_{(r)} \gets \textsc{Transform}(\theta_{\text{Gen}}^{R_i}, X^{R_i}_{(r)})$\tcp*{Domain transform}
    \textsc{Classify}($\theta_{\text{Dis}}^{R_i}$, $X^{R_i \to S}_{(r)}$)\;
    \eIf{$tr = 1 \land r = 1$}{
      $E^{R_i \to S}_{(r)} \gets \textsc{Encrypt}(X^{R_i \to S}_{(r)})$\;
      \textsc{Send}($s$, $E^{R_i \to S}_{(r)}$)\;
    }{
      \eIf{$tr = 1 \land r = 2$}{
        $\overline{E}^{\mathcal{R}\to S}_{(r-1)} \gets \textsc{Get}(s, r)$\;
        $\overline{X}^{\mathcal{R}\to S}_{(r-1)} \gets \textsc{Decrypt}(\overline{E}^{\mathcal{R}\to S}_{(r-1)})$\;
      }{
        $\overline{X}^{\mathcal{R}\to S}_{(r-1)} \gets \textsc{Get}(s, r)$\;
      }
      \textsc{Classify}($\theta_{\text{Dis}}^{R_i}$, $\overline{X}^{\mathcal{R}\to S}_{(r-1)}$)\;
      $X^{(\mathcal{R}\to S, R_i)}_{(r)} \gets \overline{X}^{\mathcal{R}\to S}_{(r-1)} + X^{R_i \to S}_{(r)} - X^{R_i \to S}_{(r-1)}$\tcp*{Masking}
      \textsc{Send}($s$, $X^{(\mathcal{R}\to S, R_i)}_{(r)}$)\;
    }
  }
}
\end{algorithm}

\begin{algorithm}[t]
\caption{The training of the FedTT framework in the server $s$}
\label{algo:2}
\KwIn{Clients $\mathcal{C}=\{c_1,\dots,c_n\}$ with source cities $\mathcal{R}=\{R_1,\dots,R_n\}$; models $\theta_{\text{Dis}}, \theta_{\text{TP}}$; local data $\mathcal{X}^S$}
\For{$tr \gets 1$ \KwTo $T$}{
  \For{$r \gets 1$ \KwTo $R$}{
    \eIf{$tr = 1 \land r = 1$}{
      $\{E^{R_1 \to S}_{(r)}, E^{R_2 \to S}_{(r)}, \dots \} \gets \textsc{Get}(\mathcal{C}, r)$\tcp*{Receive encrypted data}
      $\overline{E}^{\mathcal{R}\to S}_{(r)} \gets \sum_{i=1}^{n} E^{R_i \to S}_{(r)}$\tcp*{Aggregate encrypted data}
      \textsc{Send}($\mathcal{C}, \overline{E}^{\mathcal{R}\to S}_{(r)}$)\;
    }{
      $\{X^{(\mathcal{R}\to S, R_1)}_{(r)}, X^{(\mathcal{R}\to S, R_2)}_{(r)}, \dots \} \gets \textsc{Get}(\mathcal{C}, r)$\tcp*{Receive masked data}
      $\overline{X}^{\mathcal{R}\to S}_{(r)} \gets \sum_{i=1}^{n} X^{(\mathcal{R}\to S, R_i)}_{(r)} - (n-1)\cdot \overline{X}^{\mathcal{R}\to S}_{(r-1)}$\tcp*{Aggregate masked data}
      \textsc{Classify}($\theta_{\text{Dis}}, \overline{X}^{\mathcal{R}\to S}_{(r)}$)\;
      \textsc{Send}($\mathcal{C}, \overline{X}^{\mathcal{R}\to S}_{(r)}$)\;
    }
  }
  \textsc{Classify}($\theta_{\text{Dis}}, \mathcal{X}^S$)\tcp*{Update on server local data}
  \textsc{Prediction}($\theta_{\text{TP}}, \overline{\mathcal{X}}^{\mathcal{R}\to S}, \mathcal{X}^S$)\tcp*{Traffic prediction}
}
\end{algorithm}

\subsection{Training Process}
Before the training of the FedTT framework, clients (i.e., source cities) train the spatial view expansion model $\theta_{\textit{SV}}$ and the temporal view expansion model $\theta_{\textit{TV}}$ in the TVI module $\theta_{\textit{TVI}}$ by minimizing the loss in Eqs.~\ref{eq8} and \ref{eq12}, as shown below:
\begin{equation}
\mathop{\min}_{\theta_{\textit{TVI}}} \mathcal{L}({\theta_{\textit{TVI}}}, \mathcal{V}_\textit{SV}, \mathcal{V}_\textit{TV})=\mathop{\min}_{\theta_{\textit{SV}}} \mathcal{L}({\theta_{\textit{SV}}}, \mathcal{V}_\textit{SV})+ \mathop{\min}_{\theta_{\textit{TV}}} \mathcal{L}({\theta_{\textit{TV}}}, \mathcal{V}_\textit{TV}),
\end{equation}
where $\mathcal{V}_\textit{SV}$ and $\mathcal{V}_\textit{TV}$ are the set of traffic subviews at different times obtained by spatial view extension and temporal view enhancement, respectively. During the training of the FedTT framework, the client $c_i$ train the generator model $\theta_{\textit{Gen}}^{R_i}$ and the discriminator model $\theta_{\textit{Dis}}^{R_i}$ by minimizing the loss in Eqs.~\ref{eq28} and \ref{eq29}, as shown below:
\begin{equation}
\mathop{\min}_{\theta_{\textit{Gen}}^{R_i}} \mathcal{L}(\theta_{\textit{Gen}}^{R_i}, \theta_{\textit{Dis}},\theta_{\textit{Dis}}^{R_i}, \mathcal{X}^{R_i\to S}, \mathcal{X}^{\mathcal{R}S}, \mathcal{X}^{R_i S})+\mathop{\min}_{\theta_{\textit{Dis}}^{R_i}} \mathcal{L}({\theta_{\textit{Dis}}^{R_i}}, \mathcal{X}^{{R_i S}}),
\end{equation}
where $\mathcal{X}^{\mathcal{R}S}$ is the traffic data consisting of the aggregated data $\overline{\mathcal{X}}^{\mathcal{R} \to S}$ and traffic data $\mathcal{X}^S$ of the target city $S$, and $\mathcal{X}^{R_i S}$ is the traffic data consisting of the aggregated data $\overline{\mathcal{X}}^{\mathcal{R} \to S}$ and transformed data $\mathcal{X}^{R_i \to S}$ of the source city $R_i$. 
Besides, the server $s$ trains the discriminator model $\theta_{\textit{Dis}}$ and traffic model $\theta_{\textit{TP}}$ by minimizing the loss in Eqs.~\ref{eq21} and \ref{eq3}, as shown below:
\begin{equation}
\mathop{\min}_{\theta_{\textit{Dis}}} \mathcal{L}({\theta_{\textit{Dis}}}, {\mathcal{X}}^{\mathcal{R}S})+\mathop{\min}_{\theta_{\textit{TP}}} \mathcal{L}(\theta_{\textit{TP}}, \overline{D}^{\mathcal{R} \to S}, D^S),
\end{equation}
where $\overline{D}^{\mathcal{R} \to S}$ is the aggregated dataset whose traffic domain is transformed from source cities to the target city $S$. 



\textbf{Training Algorithm}. For convenient method reproduction, we provide detailed training Algorithms~\ref{algo:1} and~\ref{algo:2} of the FedTT framework, including the client and server.

In the client (i.e., Algorithm~\ref{algo:1}), the target city acts as the server. Before the training process, the client completes the missing traffic data through the traffic view imputation method (line 1). During each training round and each traffic data (lines 2--3), it first transforms the data from the traffic domain of the source city to that of the target city using the local generator model (line 4) and classifies the transformed data using the local discriminator model (line 5). If the training process is in the first round using the first data instance (line 6), the client encrypts the transformed data using homomorphic encryption and sends it to the server (lines 7-8). Otherwise, if the training process is in the first round using the second data instance (lines 9-10), the client gets the encrypted data and decrypts it to get the previous aggregated data (lines 11-12). For subsequent rounds or data instance, the client directly gets the previous aggregated data from the server without decryption (lines 13-14). In either case, it classifies the previous aggregated data using its local discriminator model (line 15). Then it masks the transformed data using the previous aggregated and transformed data (line 16). Finally, it sends the mask data to the server for data aggregation (lines 17).

In the server (i.e., Algorithm~\ref{algo:2}), the source cities act as the clients. During each training round and each traffic data (lines 1--2), if the training process is in the first round using the first data instance (line 3), the server gets the encrypted data from clients (line 4). Then, it aggregates them by summing up, and send the aggregated encrypted data to back to the clients for further processing (lines 5-6). For subsequent rounds or data instances (line 7), the server gets the mask data from clients (line 8). Then, it aggregates the masked data using the previous aggregated data (line 9). Next, it classifies the aggregated data using its global discriminator model and sends the aggregated data back to the clients (lines 10--11). Finally, at the end of each training round, it classifies local traffic data and performs traffic prediction using the aggregated and local traffic data (lines 12--13).

\textbf{Complexity Analysis}. We also give the complete complexity analysis for the training of the FedTT framework, i.e., Algorithms~\ref{algo:1} and \ref{algo:2}. For the client (i.e., Algorithm~\ref{algo:1}), the training complexity is $O((|\mathcal{M}^{R_i}|+ |\mathcal{M}^S|) \times (F_1 \times H)^2 \times |\mathcal{X}^{R_i}|)$ at each round. For the server (i.e., Algorithm~\ref{algo:2}), the training complexity is $O((|\mathcal{M}^S| \times (F_1 \times H)^2 + \textit{MC}(\theta_{\textit{TP}})) \times (|\mathcal{X}^S|+\sum_{i=1}^{n} |\mathcal{X}^{R_i}|))$ at each round. Here, $|\mathcal{M}^{R_i}|$ and $|\mathcal{M}^S|$ are the number of sensors in the source city $R_i$ and target city $S$, respectively. Besides, $|\mathcal{X}^{R_i}|$ and $|\mathcal{X}^S|$ are the number of traffic data in the source city $R_i$ and target city $S$, respectively. In addition, $F_1=3$ is the dimensions of traffic data features, and $H=1024$ is the hidden dimensions of the three-layer MLP model in $\theta_{\textit{Gen}}^{R_i}$ and $\theta_{\textit{Dis}}^{R_i}$. Moreover, $\textit{MC}(\theta_{\textit{TP}})$ is the model complexity of $\theta_{\textit{TP}}$ (i.e., $\theta_{\textit{DyHSL}}$).

\section{Experiment}
\vspace{-1mm}
\label{sec:exp}

\begin{table}[tb]
\caption{Statistics of Evaluated Datasets}
\vspace{-2mm}
\label{tab:datasets_transposed}
\resizebox{\linewidth}{!}{
\begin{tabular}{|c|c|c|c|c|}
\hline
{Dataset} & {\# instances} & {\# sensors} & {Interval} & {City} \\\hline
{PeMSD4}  & 16992                & 307                 & 5 mins           & San Francisco       \\\hline
{PeMSD8}  & 17856                & 170                 & 5 mins           & San Bernardino         \\\hline
{FT-AED}  & 1920                 & 196                 & 5 mins           & Nashville        \\\hline
{HK-Traffic}& 17856                & 411                 & 5 mins         & Hong Kong        \\\hline
\end{tabular}
}
\label{tab:datasets}
\vspace{-4mm}
\end{table}

\begin{table*}[]
\vspace{-2mm}
\caption{The Overall Performance Comparison between Different Methods}
\vspace{-2mm}
\resizebox{\linewidth}{!}{
\begin{threeparttable}
\begin{tabular}{|c|c|ccc|ccc|ccc|ccc|}
\hline
\multirow{2}{*}{{Metric}} & \multirow{2}{*}{{Method}} & \multicolumn{3}{c|}{{(P8, FT, HK)} $\to$ {P4$^1$}}                                                            & \multicolumn{3}{c|}{{(P4, FT, HK)} $\to$ {P8}}                                                            & \multicolumn{3}{c|}{{(P4, P8, HK)} $\to$ {FT}}                                                            & \multicolumn{3}{c|}{{(P4, P8, FT)} $\to$ {HK}}                                                           \\ \cline{3-14} 
                                 &                                  & \multicolumn{1}{c}{{flow}}  & \multicolumn{1}{c}{{speed}} & {occ}    & \multicolumn{1}{c}{{flow}}  & \multicolumn{1}{c}{{speed}} & {occ}    & \multicolumn{1}{c}{{flow}}  & \multicolumn{1}{c}{{speed}} & {occ}    & \multicolumn{1}{c}{{flow}} & \multicolumn{1}{c}{{speed}} & {occ}    \\ \hline
\multirow{9}{*}{{MAE}}    & {pFedCTP}                 & \multicolumn{1}{c}{21.24}          & \multicolumn{1}{c}{1.52}           & 0.0079          & \multicolumn{1}{c}{17.06}          & \multicolumn{1}{c}{1.22}           & 0.0072          & \multicolumn{1}{c}{13.92}          & \multicolumn{1}{c}{5.78}           & 0.0415          & \multicolumn{1}{c}{9.22}          & \multicolumn{1}{c}{1.22}           & 0.0102          \\  
                                 & {T-ISTGNN}                & \multicolumn{1}{c}{27.24}          & \multicolumn{1}{c}{2.03}           & 0.0219          & \multicolumn{1}{c}{22.75}          & \multicolumn{1}{c}{1.84}           & 0.0235          & \multicolumn{1}{c}{20.83}          & \multicolumn{1}{c}{9.69}           & 0.0571          & \multicolumn{1}{c}{9.98}          & \multicolumn{1}{c}{4.24}           & 0.0121          \\  
                                 & {TPB}                     & \multicolumn{1}{c}{21.06}          & \multicolumn{1}{c}{1.28}           & 0.0134          & \multicolumn{1}{c}{17.11}          & \multicolumn{1}{c}{1.12}           & 0.0081          & \multicolumn{1}{c}{13.03}          & \multicolumn{1}{c}{3.59}           & 0.0276          & \multicolumn{1}{c}{8.36}          & \multicolumn{1}{c}{1.52}           & 0.0092          \\  
                                 & {ST-GFSL}                 & \multicolumn{1}{c}{23.05}          & \multicolumn{1}{c}{1.47}           & 0.0161          & \multicolumn{1}{c}{19.86}          & \multicolumn{1}{c}{1.47}           & 0.0159          & \multicolumn{1}{c}{18.00}          & \multicolumn{1}{c}{5.25}           & 0.0385          & \multicolumn{1}{c}{8.42}          & \multicolumn{1}{c}{2.03}           & 0.0101          \\  
                                 & {DastNet}                 & \multicolumn{1}{c}{26.89}          & \multicolumn{1}{c}{1.54}           & 0.0165          & \multicolumn{1}{c}{19.58}          & \multicolumn{1}{c}{1.41}           & 0.0134          & \multicolumn{1}{c}{15.44}          & \multicolumn{1}{c}{4.62}           & 0.0421          & \multicolumn{1}{c}{9.09}          & \multicolumn{1}{c}{3.85}           & 0.0135          \\  
                                 & {CityTrans}               & \multicolumn{1}{c}{23.94}          & \multicolumn{1}{c}{1.38}           & 0.0119          & \multicolumn{1}{c}{18.51}          & \multicolumn{1}{c}{1.18}           & 0.0108          & \multicolumn{1}{c}{13.06}          & \multicolumn{1}{c}{3.60}           & 0.0359          & \multicolumn{1}{c}{8.78}          & \multicolumn{1}{c}{1.84}           & 0.0116          \\  
                                 & {TransGTR}                & \multicolumn{1}{c}{24.32}          & \multicolumn{1}{c}{1.39}           & 0.0135          & \multicolumn{1}{c}{19.53}          & \multicolumn{1}{c}{1.18}           & 0.0089          & \multicolumn{1}{c}{13.27}          & \multicolumn{1}{c}{4.80}           & 0.0337          & \multicolumn{1}{c}{9.09}          & \multicolumn{1}{c}{3.92}           & 0.0102          \\  
                                 & {MGAT}                    & \multicolumn{1}{c}{24.78}          & \multicolumn{1}{c}{1.58}           & 0.0195          & \multicolumn{1}{c}{20.16}          & \multicolumn{1}{c}{1.67}           & 0.0160          & \multicolumn{1}{c}{20.08}          & \multicolumn{1}{c}{8.00}           & 0.0469          & \multicolumn{1}{c}{9.14}          & \multicolumn{1}{c}{2.88}           & 0.0101          \\ \cline{2-14} 
                                 & {FedTT}                   & \multicolumn{1}{c}{\textcolor{lightblue}{16.69}} & \multicolumn{1}{c}{\textcolor{lightblue}{1.03}}  & \textcolor{lightblue}{0.0061} & \multicolumn{1}{c}{\textcolor{lightblue}{14.11}} & \multicolumn{1}{c}{\textcolor{lightblue}{0.94}}  & \textcolor{lightblue}{0.0059} & \multicolumn{1}{c}{\textcolor{lightblue}{12.10}} & \multicolumn{1}{c}{\textcolor{lightblue}{3.24}}  & \textcolor{lightblue}{0.0249} & \multicolumn{1}{c}{\textcolor{lightblue}{7.42}} & \multicolumn{1}{c}{\textcolor{lightblue}{1.05}}  & \textcolor{lightblue}{0.0087} \\ \hline
\multirow{9}{*}{{RMSE}}   & {pFedCTP}                 & \multicolumn{1}{c}{33.03}          & \multicolumn{1}{c}{3.12}           & 0.0188          & \multicolumn{1}{c}{26.19}          & \multicolumn{1}{c}{2.62}           & 0.0164          & \multicolumn{1}{c}{19.94}          & \multicolumn{1}{c}{9.84}           & 0.0756          & \multicolumn{1}{c}{13.31}         & \multicolumn{1}{c}{2.62}           & 0.0212          \\  
                                 & {T-ISTGNN}                & \multicolumn{1}{c}{35.95}          & \multicolumn{1}{c}{4.14}           & 0.0281          & \multicolumn{1}{c}{31.10}          & \multicolumn{1}{c}{3.37}           & 0.0305          & \multicolumn{1}{c}{29.42}          & \multicolumn{1}{c}{13.17}          & 0.1127          & \multicolumn{1}{c}{15.68}         & \multicolumn{1}{c}{6.31}           & 0.0230          \\  
                                 & {TPB}                     & \multicolumn{1}{c}{31.75}          & \multicolumn{1}{c}{2.31}           & 0.0201          & \multicolumn{1}{c}{26.35}          & \multicolumn{1}{c}{2.19}           & 0.0126          & \multicolumn{1}{c}{16.34}          & \multicolumn{1}{c}{6.07}           & 0.0493          & \multicolumn{1}{c}{11.89}         & \multicolumn{1}{c}{2.98}           & 0.0152          \\  
                                 & {ST-GFSL}                 & \multicolumn{1}{c}{33.65}          & \multicolumn{1}{c}{3.29}           & 0.0237          & \multicolumn{1}{c}{30.66}          & \multicolumn{1}{c}{3.12}           & 0.0260          & \multicolumn{1}{c}{22.10}          & \multicolumn{1}{c}{9.69}           & 0.0652          & \multicolumn{1}{c}{12.89}         & \multicolumn{1}{c}{4.73}           & 0.0156          \\  
                                 & {DastNet}                 & \multicolumn{1}{c}{34.96}          & \multicolumn{1}{c}{3.41}           & 0.0274          & \multicolumn{1}{c}{27.45}          & \multicolumn{1}{c}{3.10}           & 0.0299          & \multicolumn{1}{c}{22.64}          & \multicolumn{1}{c}{9.72}           & 0.0691          & \multicolumn{1}{c}{13.63}         & \multicolumn{1}{c}{5.82}           & 0.0236          \\  
                                 & {CityTrans}               & \multicolumn{1}{c}{32.04}          & \multicolumn{1}{c}{2.46}           & 0.0237          & \multicolumn{1}{c}{27.91}          & \multicolumn{1}{c}{2.20}           & 0.0226          & \multicolumn{1}{c}{18.86}          & \multicolumn{1}{c}{9.82}           & 0.0514          & \multicolumn{1}{c}{13.45}         & \multicolumn{1}{c}{4.72}           & 0.0212          \\  
                                 & {TransGTR}                & \multicolumn{1}{c}{33.66}          & \multicolumn{1}{c}{2.43}           & 0.0198          & \multicolumn{1}{c}{26.41}          & \multicolumn{1}{c}{2.27}           & 0.0147          & \multicolumn{1}{c}{17.11}          & \multicolumn{1}{c}{7.96}           & 0.0579          & \multicolumn{1}{c}{12.23}         & \multicolumn{1}{c}{6.77}           & 0.0180          \\  
                                 & {MGAT}                    & \multicolumn{1}{c}{32.85}          & \multicolumn{1}{c}{3.43}           & 0.0283          & \multicolumn{1}{c}{30.77}          & \multicolumn{1}{c}{3.20}           & 0.0262          & \multicolumn{1}{c}{24.62}          & \multicolumn{1}{c}{11.05}          & 0.1028          & \multicolumn{1}{c}{12.03}         & \multicolumn{1}{c}{5.11}           & 0.0162          \\ \cline{2-14} 
                                 & {FedTT}                   & \multicolumn{1}{c}{\textcolor{lightblue}{27.48}} & \multicolumn{1}{c}{\textcolor{lightblue}{1.93}}  & \textcolor{lightblue}{0.0166} & \multicolumn{1}{c}{\textcolor{lightblue}{24.29}} & \multicolumn{1}{c}{\textcolor{lightblue}{1.94}}  & \textcolor{lightblue}{0.0099} & \multicolumn{1}{c}{\textcolor{lightblue}{15.91}} & \multicolumn{1}{c}{\textcolor{lightblue}{5.50}}  & \textcolor{lightblue}{0.0372} & \multicolumn{1}{c}{\textcolor{lightblue}{8.57}} & \multicolumn{1}{c}{\textcolor{lightblue}{2.40}}  & \textcolor{lightblue}{0.0145} \\ \hline
\end{tabular}
\begin{tablenotes}
\item[1] {P4, P8, FT, and HK denote PeMSD4, PeMSD8, FT-AED, and HK-Traffic datasets, respectively.}
\end{tablenotes}
\end{threeparttable}
}
\label{tab:op}
\vspace{-2mm}
\end{table*}

\begin{table*}[]
\caption{The Overall Performance (MAE) Comparison when Extending Centralized Traffic Models}
\vspace{-2mm}
\resizebox{\linewidth}{!}{
\begin{threeparttable}
\begin{tabular}{|c|c|ccc|ccc|ccc|ccc|}
\hline
\multirow{2}{*}{{Model}}    & \multirow{2}{*}{{Method}} & \multicolumn{3}{c|}{(P8, FT, HK) $\to$ P4}                                                            & \multicolumn{3}{c|}{{(P4, FT, HK)} $\to$ {P8}}                                                            & \multicolumn{3}{c|}{{(P4, P8, HK)} $\to$ {FT}}                                                            & \multicolumn{3}{c|}{{(P4, P8, FT)} $\to$ {HK}}                                                             \\ \cline{3-14} 
                                   &                                  & \multicolumn{1}{c}{{flow}}  & \multicolumn{1}{c}{{speed}} & {occ}    & \multicolumn{1}{c}{{flow}}  & \multicolumn{1}{c}{{speed}} & {occ}    & \multicolumn{1}{c}{{flow}}  & \multicolumn{1}{c}{{speed}} & {occ}    & \multicolumn{1}{c}{{flow}} & \multicolumn{1}{c}{{speed}} & {occ}    \\ \hline
\multirow{2}{*}{{GRU}}      & {FTL$^1$}                     & \multicolumn{1}{c}{29.27}          & \multicolumn{1}{c}{3.39}           & 0.0282          & \multicolumn{1}{c}{23.44}          & \multicolumn{1}{c}{2.40}           & 0.0253          & \multicolumn{1}{c}{21.16}          & \multicolumn{1}{c}{12.18}          & 0.0712          & \multicolumn{1}{c}{10.11}         & \multicolumn{1}{c}{4.60}           & 0.0125          \\  
                                   & {FedTT}                   & \multicolumn{1}{c}{\textcolor{lightblue}{25.93}} & \multicolumn{1}{c}{\textcolor{lightblue}{2.24}}  & \textcolor{lightblue}{0.0220} & \multicolumn{1}{c}{\textcolor{lightblue}{20.73}} & \multicolumn{1}{c}{\textcolor{lightblue}{2.21}}           & \textcolor{lightblue}{0.0213} & \multicolumn{1}{c}{\textcolor{lightblue}{17.34}} & \multicolumn{1}{c}{\textcolor{lightblue}{5.67}}  & \textcolor{lightblue}{0.0401} & \multicolumn{1}{c}{\textcolor{lightblue}{9.33}} & \multicolumn{1}{c}{\textcolor{lightblue}{2.86}}  & \textcolor{lightblue}{0.0101} \\ \hline
\multirow{2}{*}{{CNN}}      & {FTL}                     & \multicolumn{1}{c}{31.46}          & \multicolumn{1}{c}{4.55}           & 0.0317          & \multicolumn{1}{c}{27.60}          & \multicolumn{1}{c}{3.27}           & 0.0267          & \multicolumn{1}{c}{24.55}          & \multicolumn{1}{c}{9.05}           & 0.0803          & \multicolumn{1}{c}{9.74}          & \multicolumn{1}{c}{5.92}           & 0.0169          \\  
                                   & {FedTT}                   & \multicolumn{1}{c}{\textcolor{lightblue}{26.82}} & \multicolumn{1}{c}{\textcolor{lightblue}{2.84}}  & \textcolor{lightblue}{0.0274} & \multicolumn{1}{c}{\textcolor{lightblue}{22.20}} & \multicolumn{1}{c}{\textcolor{lightblue}{2.41}}  & \textcolor{lightblue}{0.0217} & \multicolumn{1}{c}{\textcolor{lightblue}{17.44}} & \multicolumn{1}{c}{\textcolor{lightblue}{6.27}}  & \textcolor{lightblue}{0.0472} & \multicolumn{1}{c}{\textcolor{lightblue}{9.24}} & \multicolumn{1}{c}{\textcolor{lightblue}{3.92}}  & \textcolor{lightblue}{0.0113} \\ \hline
\multirow{2}{*}{{MLP}}      & {FTL}                     & \multicolumn{1}{c}{34.01}          & \multicolumn{1}{c}{3.66}           & 0.0276          & \multicolumn{1}{c}{30.24}          & \multicolumn{1}{c}{2.88}           & 0.0246          & \multicolumn{1}{c}{22.66}          & \multicolumn{1}{c}{14.43}          & 0.0743          & \multicolumn{1}{c}{10.87}         & \multicolumn{1}{c}{5.23}           & 0.0146          \\  
                                   & {FedTT}                   & \multicolumn{1}{c}{\textcolor{lightblue}{28.08}} & \multicolumn{1}{c}{\textcolor{lightblue}{2.17}}  & \textcolor{lightblue}{0.0250} & \multicolumn{1}{c}{\textcolor{lightblue}{23.79}} & \multicolumn{1}{c}{\textcolor{lightblue}{2.40}}  & \textcolor{lightblue}{0.0212} & \multicolumn{1}{c}{\textcolor{lightblue}{17.66}} & \multicolumn{1}{c}{\textcolor{lightblue}{7.35}}  & \textcolor{lightblue}{0.0480} & \multicolumn{1}{c}{\textcolor{lightblue}{9.68}} & \multicolumn{1}{c}{\textcolor{lightblue}{3.27}}  & \textcolor{lightblue}{0.0102} \\ \hline
\multirow{2}{*}{{ST-SSL}}   & {FTL}                     & \multicolumn{1}{c}{26.76}          & \multicolumn{1}{c}{2.26}           & 0.0176          & \multicolumn{1}{c}{20.06}          & \multicolumn{1}{c}{1.88}           & 0.0226          & \multicolumn{1}{c}{19.43}          & \multicolumn{1}{c}{7.78}           & 0.0605          & \multicolumn{1}{c}{9.43}          & \multicolumn{1}{c}{4.36}           & 0.0117          \\  
                                   & {FedTT}                   & \multicolumn{1}{c}{\textcolor{lightblue}{22.28}} & \multicolumn{1}{c}{\textcolor{lightblue}{1.34}}  & \textcolor{lightblue}{0.0096} & \multicolumn{1}{c}{\textcolor{lightblue}{17.14}} & \multicolumn{1}{c}{\textcolor{lightblue}{1.27}}  & \textcolor{lightblue}{0.0114} & \multicolumn{1}{c}{\textcolor{lightblue}{13.38}} & \multicolumn{1}{c}{\textcolor{lightblue}{4.88}}  & \textcolor{lightblue}{0.0400} & \multicolumn{1}{c}{\textcolor{lightblue}{8.76}} & \multicolumn{1}{c}{\textcolor{lightblue}{1.65}}  & \textcolor{lightblue}{0.0097} \\ \hline
\multirow{2}{*}{{DyHSL}}    & {FTL}                     & \multicolumn{1}{c}{18.61}          & \multicolumn{1}{c}{1.39}           & 0.0131          & \multicolumn{1}{c}{16.71}          & \multicolumn{1}{c}{1.40}           & 0.0144          & \multicolumn{1}{c}{16.96}          & \multicolumn{1}{c}{6.04}           & 0.0324          & \multicolumn{1}{c}{8.63}          & \multicolumn{1}{c}{2.97}           & 0.0103          \\  
                                   & {FedTT}                   & \multicolumn{1}{c}{\textcolor{lightblue}{16.69}} & \multicolumn{1}{c}{\textcolor{lightblue}{1.03}}  & \textcolor{lightblue}{0.0061} & \multicolumn{1}{c}{\textcolor{lightblue}{14.11}} & \multicolumn{1}{c}{\textcolor{lightblue}{0.94}}  & \textcolor{lightblue}{0.0059} & \multicolumn{1}{c}{\textcolor{lightblue}{12.10}} & \multicolumn{1}{c}{\textcolor{lightblue}{3.24}}  & \textcolor{lightblue}{0.0249} & \multicolumn{1}{c}{\textcolor{lightblue}{7.42}} & \multicolumn{1}{c}{\textcolor{lightblue}{1.05}}  & \textcolor{lightblue}{0.0087} \\ \hline
\multirow{2}{*}{{PDFormer}} & {FTL}                     & \multicolumn{1}{c}{26.99}          & \multicolumn{1}{c}{2.31}           & 0.0194          & \multicolumn{1}{c}{22.85}          & \multicolumn{1}{c}{1.80}           & 0.0232          & \multicolumn{1}{c}{17.92}          & \multicolumn{1}{c}{6.57}           & 0.0433          & \multicolumn{1}{c}{9.17}          & \multicolumn{1}{c}{3.29}           & 0.0108          \\  
                                   & {FedTT}                   & \multicolumn{1}{c}{\textcolor{lightblue}{22.05}} & \multicolumn{1}{c}{\textcolor{lightblue}{1.43}}  & \textcolor{lightblue}{0.0125} & \multicolumn{1}{c}{\textcolor{lightblue}{17.67}} & \multicolumn{1}{c}{\textcolor{lightblue}{1.36}}  & \textcolor{lightblue}{0.0127} & \multicolumn{1}{c}{\textcolor{lightblue}{13.09}} & \multicolumn{1}{c}{\textcolor{lightblue}{3.53}}  & \textcolor{lightblue}{0.0314} & \multicolumn{1}{c}{\textcolor{lightblue}{8.22}} & \multicolumn{1}{c}{\textcolor{lightblue}{1.22}}  & \textcolor{lightblue}{0.0091} \\ \hline
\end{tabular}
\begin{tablenotes}
\item[1] {FTL refers to the two-stage method used in existing FTT methods.}
\end{tablenotes}
\end{threeparttable}
}
\label{tab:opm}
\vspace{-4mm}
\end{table*}

We use the below Research Questions (RQs) to guide experiments.\\
\textit{\textbf{RQ1:} How does FedTT perform compared to state-of-the-art methods?}\\
\textit{\textbf{RQ2:} Can the proposed FedTT framework support various centralized traffic models for traffic prediction tasks?}\\
\textit{\textbf{RQ3:} How does FedTT perform in terms of the training efficiency?}\\
\textit{\textbf{RQ4:} How scalable is the proposed FedTT framework in different scales of the traffic dataset in the data-scarce target city?}\\
\textit{\textbf{RQ5:} Can FedTT handle the long-term traffic prediction tasks?}\\
\textit{\textbf{RQ6:} How does each module (i.e., TVI, TDA, TST, and FPT) of FedTT affect the performance of traffic knowledge transfer?}\\
\textit{\textbf{RQ7:} Is FedTT sensitive to hyperparameter (i.e., $\lambda_1$ and $\lambda_2$) settings?}\\

\vspace{-4mm}
\subsection{Experimental Settings}
\vspace{-1mm}
\subsubsection{\textbf{Datasets.}}
We use four datasets to evaluate the proposed FedTT framework in experiments, which are widely used in traffic prediction tasks~\cite{zys23,jjh23,jjw23}, as shown in Table~\ref{tab:datasets}.
Specifically, PeMSD4 (\textbf{P4})~\cite{pems}, PeMSD8 (\textbf{P8})~\cite{pems}, FT-AED (\textbf{FT})~\cite{ftaed}, and HK-Traffic (\textbf{HK})~\cite{hk} were collected in the San Francisco, San Bernardino, Nashville, and Hong Kong, respectively. Among them, three datasets are considered as three source cities, and one dataset serves as the target city, leading to four combinations: (P8, FT, HK) $\to$ P4, (P4, FT, HK) $\to$ P8, (P4, P8, HK) $\to$ FT, and (P4, P8, FT) $\to$ HK. Besides, we select traffic flow, speed, and occupancy prediction tasks for experiments, which are also widely studied in the community~\cite{zys23,jjh23,jjw23}.

\vspace{-2mm}
\subsubsection{\textbf{Baselines.}}
To answer \textbf{RQ1}, we first compare the proposed FedTT framework with two SOTA FTT methods, as shown below.
\begin{itemize}[leftmargin=*] 
    \item \textbf{T-ISTGNN~\cite{QiWBLYA23}.} It designs a spatio-temporal GNN-based approach with an inductive mode for cross-region traffic prediction.
    \item \textbf{pFedCTP~\cite{Zhang00XLC24}.} It designs an ST-Net for privacy-preserving and cross-city traffic prediction with personalized federated learning.
\end{itemize}
Besides, we choose three SOTA MTT methods extended for FTT.
\begin{itemize}[leftmargin=*] 
    \item \textbf{TPB~\cite{LiuZY23}.} It utilizes a traffic patch encoder to create a traffic pattern bank for the cross-city few-shot traffic knowledge transfer.
    \item \textbf{ST-GFSL~\cite{0005G0YFW22}.} It transfers traffic knowledge through model parameter matching to retrieve similar spatio-temporal features.
    \item \textbf{DastNet~\cite{TangQCLWM22}.} It employs graph learning and domain adaptation to create domain-invariant node embeddings for the traffic data.
\end{itemize}
Moreover, we choose three SOTA STT methods extended for FTT.
\begin{itemize}[leftmargin=*] 
    \item \textbf{CityTrans~\cite{OuyangYZZWH24}.} It proposes a domain adversarial model with knowledge transfer for spatio-temporal prediction across cities.
    \item \textbf{TransGTR~\cite{JinCY23}.} It leverages adaptive spatio-temporal knowledge and domain-invariant features for TP in data-scarce cities.
    \item \textbf{MGAT~\cite{MoG23}.} It extracts multi-granular regional features from source cities to enhance the effectiveness of knowledge transfer.
\end{itemize}
To answer \textbf{RQ2}, we extend three classic (GRU~\cite{ChoMGBBSB14}, CNN~\cite{LeCunBDHHHJ89}, and MLP~\cite{rumelhart1986}) and SOTA centralized traffic models in FedTT and the two-stage method used in existing FTT methods (referred as FTL).
\begin{itemize}[leftmargin=*] 
    \item \textbf{ST-SSL~\cite{jjh23}.} It models traffic data at attribute and structure levels for spatial and temporal heterogeneous-aware traffic prediction.
    \item \textbf{DyHSL~\cite{zys23}.} It leverages hypergraph structure information to extract dynamic and high-order relations of traffic road networks.
    \item \textbf{PDFormer~\cite{jjw23}.} It introduces self-attention and feature transformation for dynamic and flow-delay-aware traffic prediction.
\end{itemize}
\begin{figure}[t]
    \centering
    \vspace{-2mm}
    \includegraphics[width=0.49\textwidth]{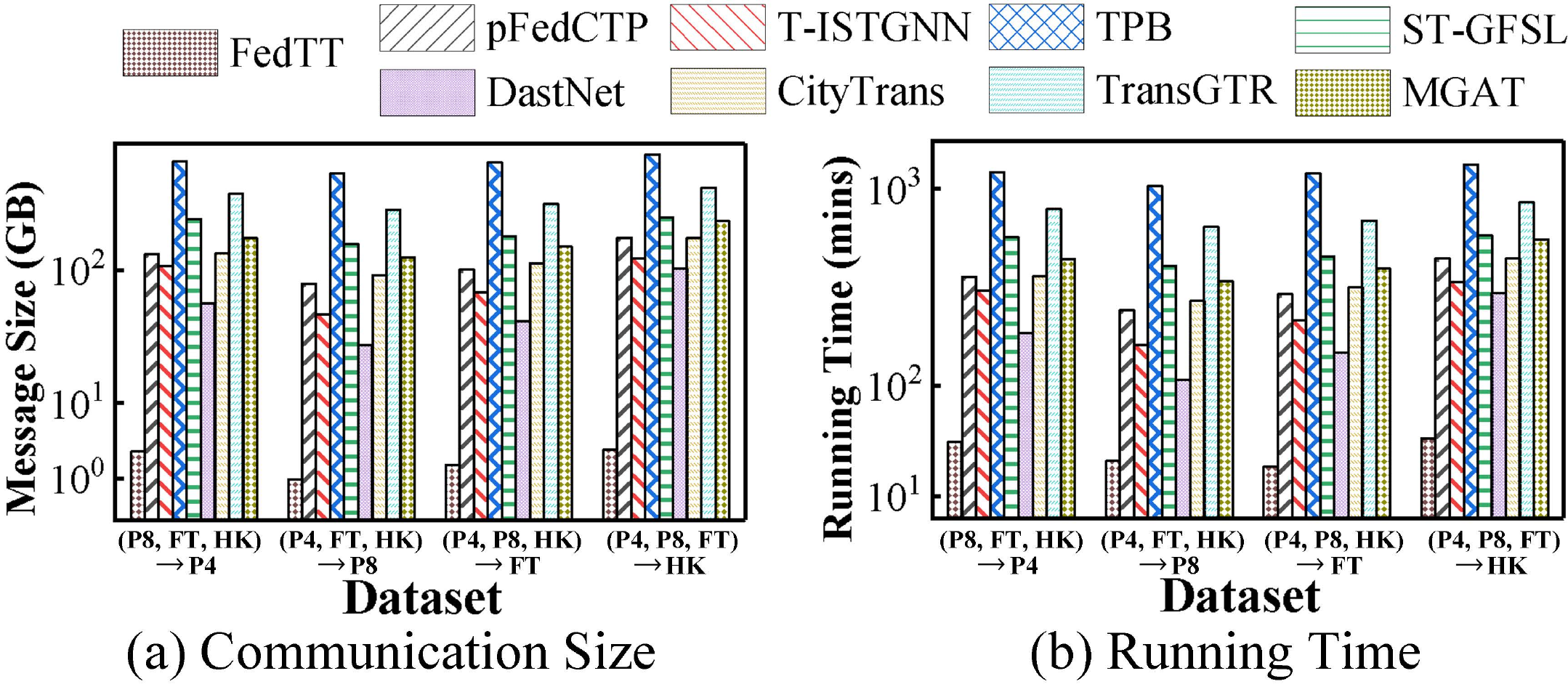}
    \vspace{-6mm}
    \caption{Training Efficiency Study of Different Methods}
    \label{fig:efficiency}
    \vspace{-2mm}
\end{figure}
\begin{figure}[t]
    \centering
    \includegraphics[width=0.49\textwidth]{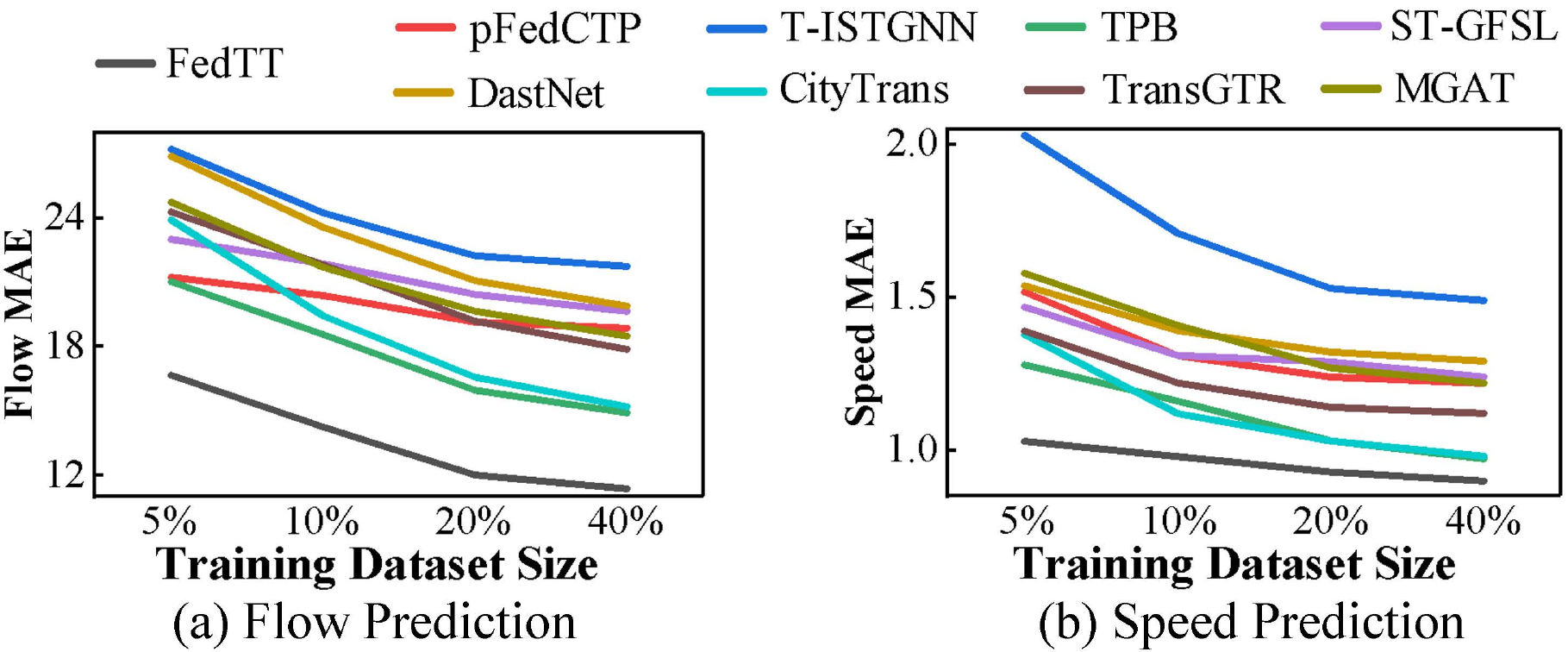}
    \vspace{-6mm}
    \caption{Model Scalability of FedTT}
    \label{fig:scala}
    \vspace{-4mm}
\end{figure}
Additionally, we choose three SOTA data imputation methods and replace the TVI module of the FedTT framework to evaluate its effects on the completion and prediction of the missing traffic data.
\begin{itemize}[leftmargin=*] 
    \item \textbf{LATC~\cite{xyc24}.} It integrates temporal variation as a regularization term to accurately impute missing spatio-temporal traffic data.
    \item \textbf{GCASTN~\cite{wcp23}.} It uses self-supervised learning and a missing-aware attention mechanism to impute the missing traffic data.
    \item \textbf{Nuhuo~\cite{hty24}.} It uses graph neural networks and self-supervised learning to accurately estimate missing traffic speed histograms.
\end{itemize}

\subsubsection{\textbf{Evaluation Metrics.}}
We use Mean Absolute Error (MAE), Root Mean Square Error (RMSE), communication size (GB), and running time (minutes) as evaluation metrics in experiments, where MAE and RMSE are used to evaluate the effectiveness, while communication size and running time are used to evaluate the efficiency. 

\subsubsection{\textbf{Implementation.}} 
All baselines run under their optimal settings. FedTT can protect traffic data privacy with the TST module, while other baselines entail a risk of privacy leakage in the data alignment and model aggregation. To protect traffic data privacy in FTT, one alternative approach for baselines is to employ Homomorphic Encryption (HE)~\cite{rlr78} and Differential Privacy (DP)~\cite{cd06}. Specifically, DP is used to perturb traffic data from other cities if the client needs to use them for data alignment in MTT and STT baselines, and HE is used to encrypt uploaded models for model aggregation across all baselines. Therefore, to safeguard traffic data privacy and ensure fairness in experiments, we extend baselines combined with this optional approach in FTT. Besides, We take 60 minutes (12-time steps) of historical data as input and output the traffic prediction in the next 15 minutes (3-time steps). In addition, we use 5\% train data, 10\% validation data, and 10\% test data of the target city in experiments. Moreover, all experiments are conducted in the federation with four nodes, one as a server and the other three nodes as clients, each equipped with two Intel Xeon CPU E5-2650 12-core processors and two NVIDIA GeForce RTX 3090.

\begin{figure}[t]
    \centering
    \vspace{-2.5mm}
    \includegraphics[width=0.49\textwidth]{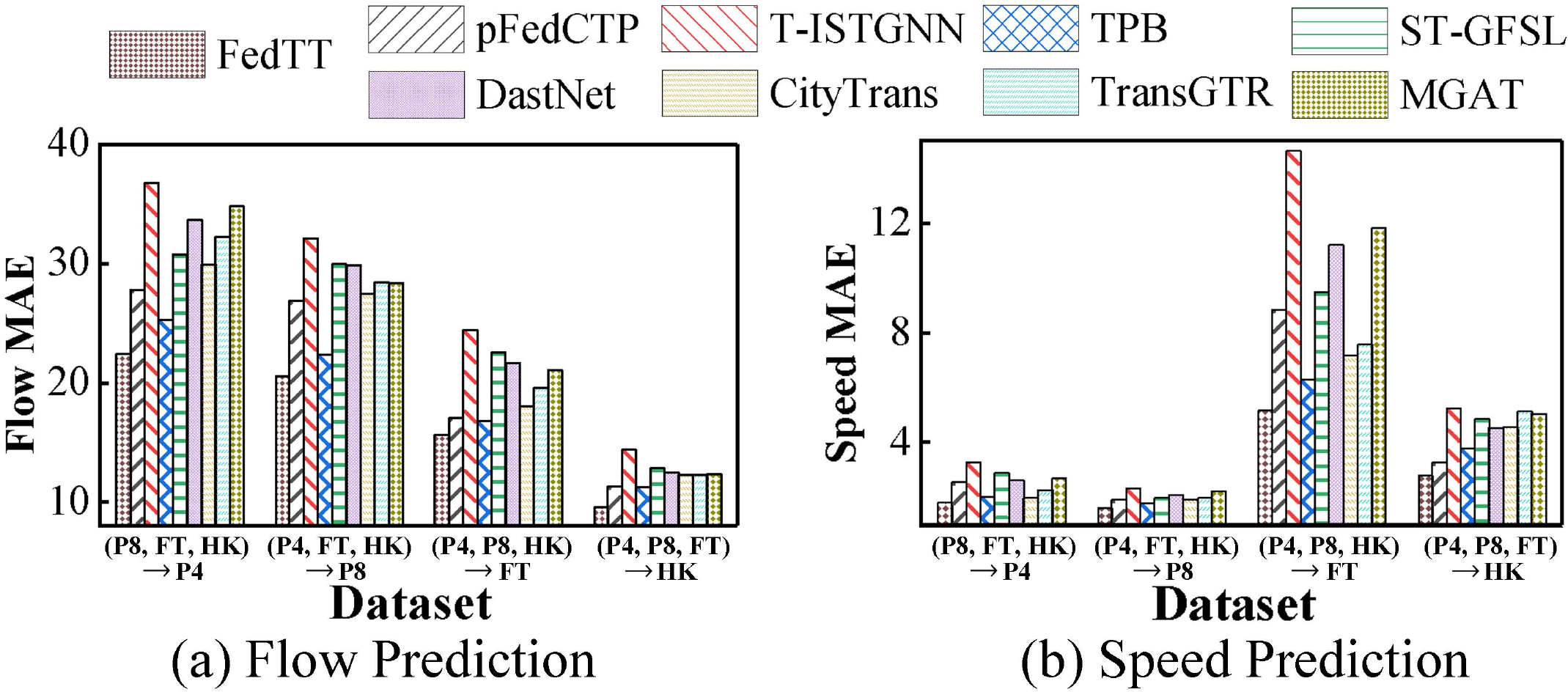}
    \vspace{-6mm}
    \caption{The Performance in Long-Term Traffic Prediction}
    \vspace{-4mm}
    \label{fig:long}
\end{figure}

\begin{figure*}[t]
    \centering
    \vspace{-2mm}
    \includegraphics[width=\textwidth]{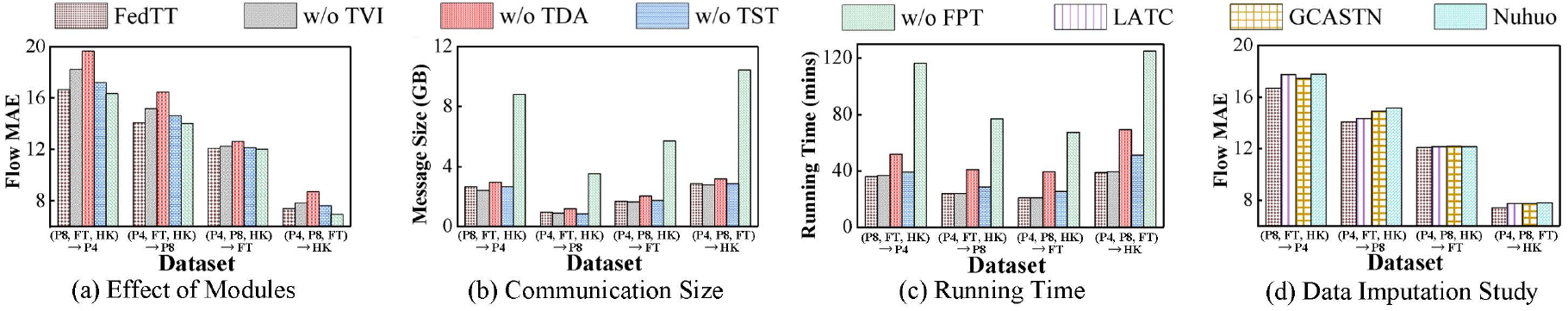}
    \vspace{-6mm}
    \caption{Ablation Study of FedTT}
    \label{fig:ab}
    \vspace{-4mm}
\end{figure*}

\begin{figure}[t]
    \centering
    \includegraphics[width=0.49\textwidth]{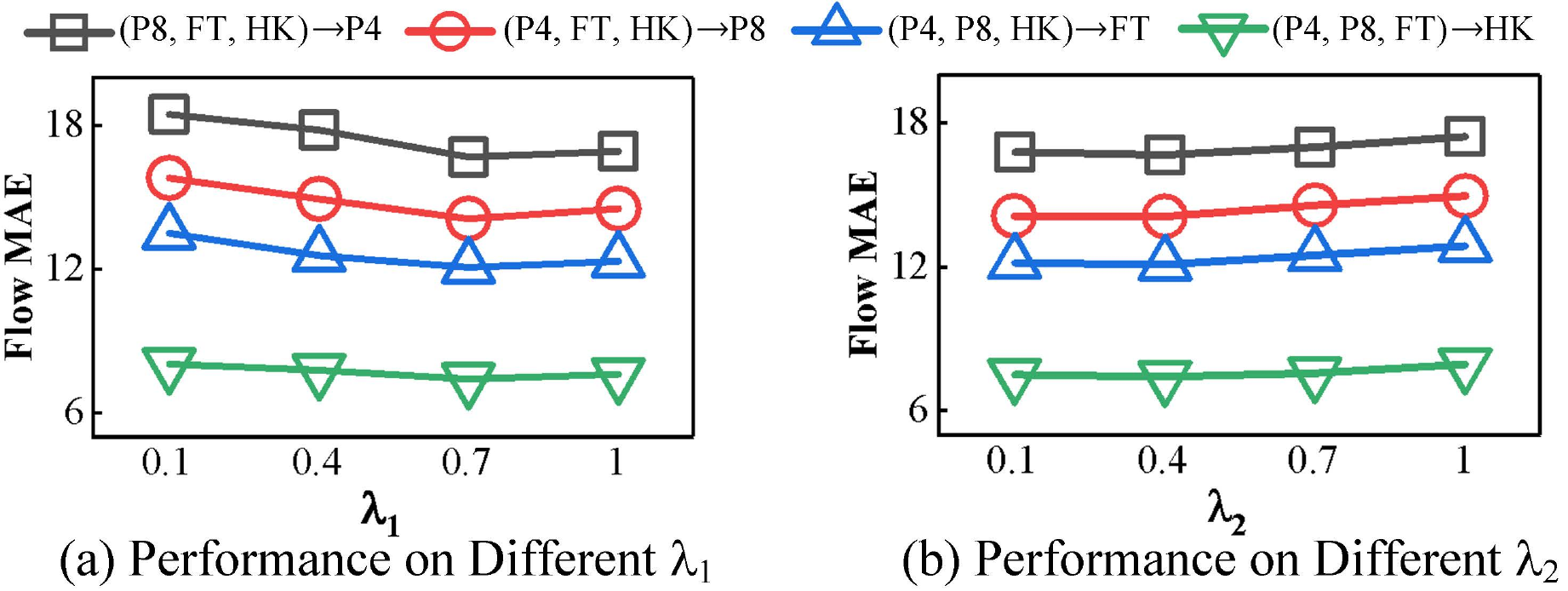}
    \vspace{-6mm}
    \caption{Parameter Sensitivity of FedTT}
    \label{fig:para}
    \vspace{-4mm}
\end{figure}

\vspace{-1mm}
\subsection{Overall Performance (RQ1 \& RQ2)}
\label{exp:op}
Table~\ref{tab:op} shows the overall performance of different methods on traffic flow, speed, and occupancy ("occ" for short) predictions, where the best results are shown in blue. Here, the DyHSL model is implemented in FedTT as it performs the best on all datasets in centralized traffic models (i.e., in Table~\ref{tab:opm}). As observed, FedTT achieves the best performance on different datasets and traffic prediction tasks compared to other methods, showing its effectiveness of traffic knowledge transfer in FTT, i.e., the gains range from \textbf{5.43\% to 22.78\%} in MAE and \textbf{7.25\% to 50.79\%} in RMSE.

Table~\ref{tab:opm} shows the overall performance when extending existing centralized traffic models in FTT using FedTT and FTL methods with MAE. As observed, all centralized traffic models extended in FedTT achieve the best performance compared to those extended in FTL, also showing its effectiveness of traffic knowledge transfer in FTT, i.e., the gains range from \textbf{5.13\% to 64.65\%}. Note that DyHSL has the best performance in centralized traffic models and is implemented in FedTT as the default model in other experiments.

\vspace{-1mm}
\subsection{Training Efficiency (RQ3)}
Fig.~\ref{fig:efficiency} shows the communication size (GB) and running time (minutes) of different methods on traffic flow prediction. As observed, the FedTT framework has the least communication size and running time compared to other methods, i.e., with communication overhead reduced by \textbf{90\%} and running time reduced by \textbf{1 to 2 orders of magnitude}, showing its superior efficiency of traffic knowledge transfer in FTT. This is because FedTT securely transmits and aggregates the traffic domain-transformed data using the TST module with relatively small computation and communication overheads, compared to other methods that employ the HE method for model secure aggregation in FTT. Besides, FedTT utilizes the FPT module to reduce data transmission and train models in parallel, significantly improving the training efficiency in FTT.

\vspace{-1mm}
\subsection{Model Scalability (RQ4)}
Fig.~\ref{fig:scala} illustrates the traffic flow and speed prediction performance of the FedTT framework across different sizes of training data in the target city, ranging from 5\% to 40\%, using the (P8, FT, HK) $\to$ P4 dataset with MAE. As observed, FedTT achieves the best performance in different-scale datasets with \textbf{7.22\% to 24.69\%} MAE less than other methods, indicating its superior scalability in FTT. Besides, as the size of the training data increases, all methods exhibit improved performance. This is because more training data enhances the model learning capability on the target city's traffic pattern.

\vspace{-1mm}
\subsection{Long-Term Traffic Prediction (RQ5)}
Fig.~\ref{fig:long} shows the performance of different methods in long-term traffic prediction in the next 60 minutes (12-time steps) with traffic flow and speed MAE. As observed, the FedTT framework achieves the best performance in FTT compared to other methods, showing its effectiveness of traffic knowledge transfer in long-term traffic prediction, i.e., the gains range from \textbf{6.86\% to 17.30\%}. Therefore, FedTT can handle both long-term and short-term traffic prediction (i.e., Table~\ref{tab:op}), showing its general advantages in FTT. 

\vspace{-1mm}
\subsection{Ablation Study (RQ6)}
\label{exp:ab}
Fig.~\ref{fig:ab} shows the ablation study of the FedTT framework, where we systematically removed the module (i.e., TVI, TDA, TST, and FPT) of FedTT one at a time, namely FedTT without TVI (w/o TVI), FedTT without TDA (w/o TDA), FedTT without TST (w/o TST), and FedTT without FPT (w/o FPT). 
First, when the TVI module is absent, the model MAE increases by \textbf{1.49\% to 9.23\%}, underscoring its pivotal role as an effective way to complete and predict the missing traffic data. Besides, the training of the TVI model is completed before the training of FedTT as it only needs to be conducted within each source city, thus not increasing communication overhead or running time during traffic model training. Additionally, compared to other data imputation methods (i.e., LATC, GCASTN, and Nuhuo), FedTT with TVI achieves better performance, showing its effectiveness in the completion and prediction of the traffic data.
Second, when the TDA module is removed, the model MAE increases by \textbf{4.46\% to 17.86\%}, which demonstrates its effectiveness in addressing the impact of traffic data distribution differences on model performance. Besides, communication overhead and running time of FedTT slightly increase compared to w/o TDA.
Third, the model MAE of FedTT decreases \textbf{0.66\% to 3.76\%} compared to w/o TST as TST uses the averaged source data for model training in the target city, which reduces the influence of specific traffic patterns of the source city on the traffic model. Besides, the communication overhead and running time of FedTT compared to w/o TST do not change as TST is a lightweight module for federated secure aggregation.
Fourth, the model MAE of FedTT does not change significantly compared to w/o FPT, but its communication overhead and running time are significantly reduced with almost 3 times less as FPT reduces data transmission and improves training efficiency.

\vspace{-1mm}
\subsection{Parameter Sensitivity (RQ7)}
\label{sec:para}
Fig.~\ref{fig:para} shows the performance of the FedTT framework with different hyperparameter settings (i.e., $\lambda_1$ and $\lambda_2$) on traffic flow prediction with MAE. First, the suggestion and optimum value of $\lambda_1$ is 0.7. As $\lambda_1$ increases, the generator model tends to generate the data that can "trick" the server discriminator model rather than generating the high-quality traffic domain transformed data, resulting in higher MAE. As $\lambda_1$ decreases, the server discriminator model loses its ability to effectively guide the generator model in generating traffic domain transformed data, resulting in higher MAE. Second, the suggestion and optimum value of $\lambda_2$ is 0.4. As $\lambda_2$ increases, the generator model tends to generate the data with a traffic domain that deviates significantly from that of the target city, resulting in higher MAE. As $\lambda_2$ decreases, the generator model generates the data with a more local-specific traffic pattern, which hinders the model from effectively learning the traffic patterns of the target city, resulting in higher MAE. Overall, FedTT has the best performance in all hyperparameter settings when $\lambda_1=0.7$ and $\lambda_2=0.4$, which are used in FedTT as the default values in other experiments.

\vspace{-1mm}
\section{Conclusion}
\label{sec:conclusion}

In this paper, we propose FedTT, an effective, efficient, privacy-aware cross-city traffic knowledge transfer framework, which transforms the traffic domain of data from source cities and trains the traffic model on transformed data in the target city. First, we design a traffic view imputation method to complete and predict missing traffic data by capturing spatio-temporal dependencies of traffic view. Second, we utilize a traffic domain adapter to uniformly transform the traffic data from the traffic domains of source cities into that of the target city. Third, we propose a traffic secret transmission method to securely transmit and aggregate the transformed data using a lightweight traffic secret aggregation method. Fourth, we introduce a federated parallel training method to enable the simultaneous training of modules. Experiments using 4 datasets on three mainstream traffic prediction tasks demonstrate the superiority of the framework. In future work, we aim to extend FedTT to support a broader range of spatio-temporal prediction tasks. 

\balance
\bibliographystyle{ACM-Reference-Format}
\bibliography{ref}

\end{document}